\documentclass[lettersize,journal]{IEEEtran}
\usepackage{amsmath,amsfonts}
\usepackage[ruled,linesnumbered]{algorithm2e}
\usepackage{array}
\newcolumntype{P}[1]{>{\centering\arraybackslash}p{#1}}
\usepackage[caption=false,font=normalsize,labelfont=sf,textfont=sf]{subfig}
\usepackage{textcomp}
\usepackage{stfloats}
\usepackage{url}
\usepackage{verbatim}
\usepackage{graphicx}
\usepackage{cite}
\usepackage{bbm}
\usepackage{dsfont}
\newcommand{\zdlmathds}[1]{\text{\usefont{U}{dsrom}{m}{n}#1}}
\usepackage{algpseudocode}
\usepackage{booktabs}
\usepackage{multirow}
\usepackage{makecell}
\usepackage{color}

\usepackage[flushleft]{threeparttable} 
\usepackage[caption=false,font=normalsize,labelfont=sf,textfont=sf]{subfig}
\usepackage{textcomp}
\usepackage{stfloats}
\usepackage{url}
\usepackage{verbatim}

\def\BibTeX{{\rm B\kern-.05em{\sc i\kern-.025em b}\kern-.08em
		T\kern-.1667em\lower.7ex\hbox{E}\kern-.125emX}}
\usepackage{balance}
\begin{document}
	\bstctlcite{IEEEexample:BSTcontrol} 
	\title{BC-GAN: A Generative Adversarial Network for Synthesizing a Batch of Collocated Clothing}
	\author{Dongliang Zhou, Haijun Zhang, Jianghong Ma, and Jianyang Shi
		\thanks{This work was supported in part by the National Natural Science Foundation of China under Grant no. 61972112, no. 61832004, and no. 62202122, the Guangdong Basic and Applied Basic Research Foundation under Grant no. 2021B1515020088, the Shenzhen Science and Technology Program under Grant no. JCYJ20210324131203009, and the HITSZ-J\&A Joint Laboratory of Digital Design and Intelligent Fabrication under Grant no. HITSZ-J\&A-2021A01.}
		\thanks{Authors are with the Department of Computer Science, Harbin Institute of
			Technology, Shenzhen, Xili University Town, Shenzhen 518055, P. R. China
			(Corresponding author: Haijun Zhang, Email: hjzhang@hit.edu.cn).}}
	
	\markboth{Journal of \LaTeX\ Class Files,~Vol.~18, No.~9, September~2020}%
	{How to Use the IEEEtran \LaTeX \ Templates}
	
	\maketitle
	
	\begin{abstract}
		Collocated clothing synthesis using generative networks has become an emerging topic in the field of fashion intelligence, as it has significant potential economic value to increase revenue in the fashion industry. In previous studies, several works have attempted to synthesize visually-collocated clothing based on a given clothing item using generative adversarial networks (GANs) with promising results. These works, however, can only accomplish the synthesis of one collocated clothing item each time. Nevertheless, users may require different clothing items to meet their multiple choices due to their personal tastes and different dressing scenarios. To address this limitation, we introduce a novel batch clothing generation framework, named \textit{BC-GAN}, which is able to synthesize multiple visually-collocated clothing images simultaneously. In particular, to further improve the fashion compatibility of synthetic results, BC-GAN proposes a new fashion compatibility discriminator in a contrastive learning perspective by fully exploiting the collocation relationship among all clothing items. Our model was examined in a large-scale dataset with compatible outfits constructed by ourselves. Extensive experiment results confirmed the effectiveness of our proposed BC-GAN in comparison to state-of-the-art methods in terms of diversity, visual authenticity, and fashion compatibility.
	\end{abstract}
	
	\begin{IEEEkeywords}
		batch generation, compatibility learning, clothing synthesis, fashion intelligence, image-to-image translation.
	\end{IEEEkeywords}
	\vspace{-0.4cm}
	
	\section{Introduction}
\begin{figure}[t]
		\centering
		\includegraphics[width=0.48\textwidth]{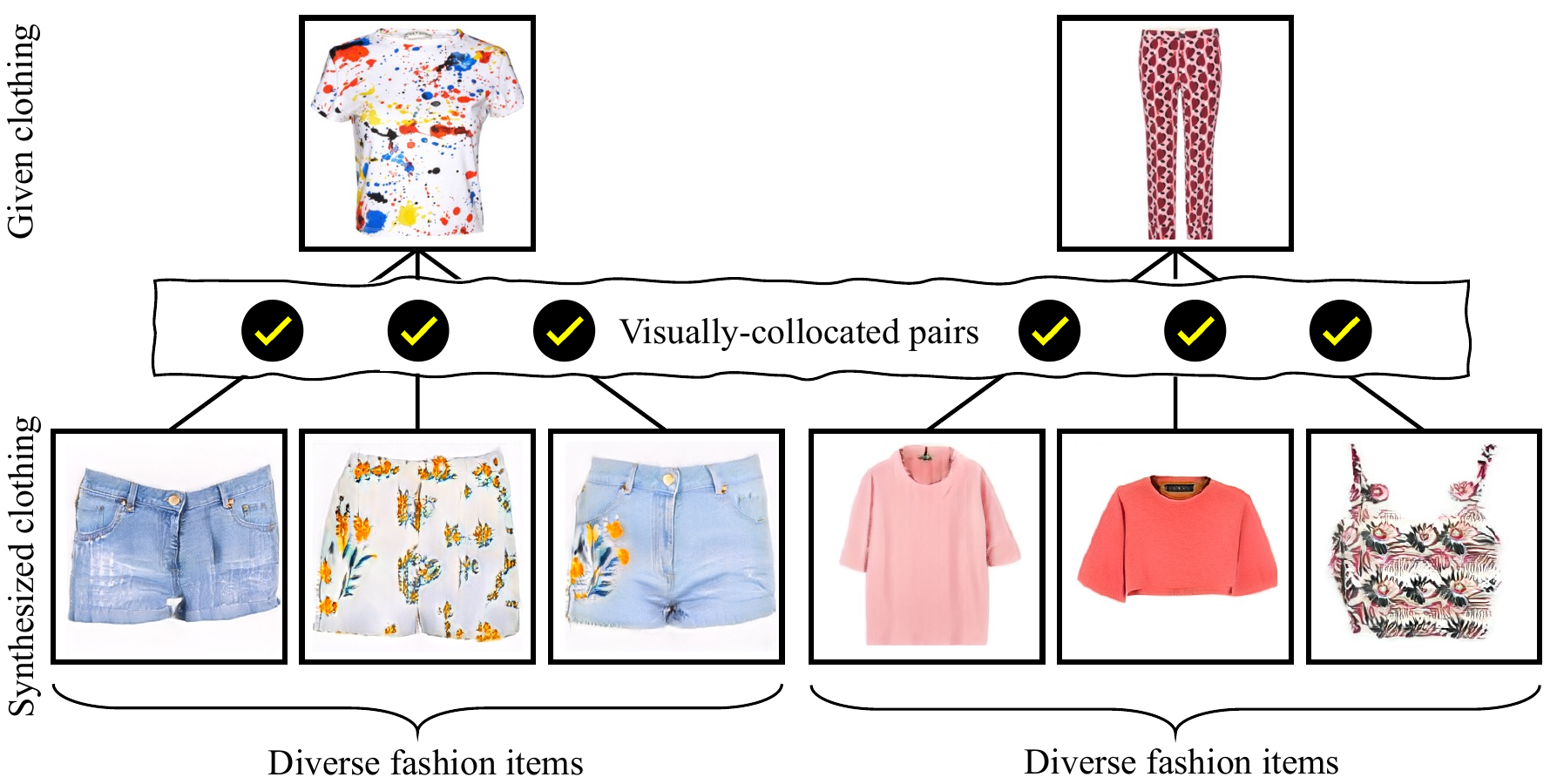}
		\caption{Multiple visually-collocated fashion items synthesized by our BC-GAN.}
		\label{cover}
		\vspace{-0.4cm}
	\end{figure}

	\IEEEPARstart{R}{ecommending} visually-collocated clothing is of significant value to the fashion industry and to promoting the business of e-commerce entities. For example, in recent years, there have been several proliferating platforms such as iFashion, ZOZO, and Xiaohongshu, which have launched their fashion matching applications, aiming to improve the turnover of clothing. As a result, this has also attracted considerable attention from the academic community. Researchers have carried out a series of studies on learning fashion compatibility, usually leveraging algorithms to determine whether a composed outfit is collocated or not \cite{mcauley2015image,veit2015learning,han2017learning,vasileva2018learning,cui2019dressing,li2020hierarchical}. The focus of these works, however, mainly lies in finding how to recommend appropriate fashion items that are compatible with the ones bought by customers, conditioned on the existence of the recommended fashion items in a database. For learning fashion compatibility, the flip side of the coin is to help fashion designers create new collocated apparel. Since generative adversarial networks (GANs) \cite{goodfellow2014generative,yang2021spatiotemporal,yan2022texture,xia2021local,zhu2021learning,zhu2020knowledge,zhu2019deep} have gradually become prominent in image synthesis, they have the potential to provide design inspirations for fashion designers through generating large pools of new garments which are compatible with existing ones \cite{liu2019toward,liu2019collocating,yu2019personalized,outfitgan,coutfitgan}. Specifically, for generating collocated clothing generation, Liu \textit{et al.} \cite{liu2019toward} first used a GAN framework, named Attribute-GAN, to assist designers in creating new apparel from a fashion compatibility learning perspective. In their model, a compatible upper clothing item can be synthesized based on a given lower clothing item, or vice versa. Liu \textit{et al.} \cite{liu2019collocating} then extended Attribute-GAN with a multi-discriminator architecture to improve the synthesis quality. The above studies, however, overlooked the personalized needs of users for clothing collocation. To address this limitation, Yu \textit{et al.} \cite{yu2019personalized} developed a collocated clothing generation framework for a personalized design. Their method works on the collocated clothing generation tasks between upper and lower clothing domains with user preference features. Although these works focused on only learning the mapping between upper and lower clothing domains, they shed light on the feasibility of collocated clothing synthesis toward intelligent design and even paved the way to outfit generation using GANs \cite{outfitgan,coutfitgan}. Nevertheless, three main drawbacks of these studies remain to be solved: (1) they can only synthesize a single image of compatible clothing at a time based on a given clothing item; (2) these methods mainly rely on general image-to-image (I2I) translation frameworks, which may be ineffective in rendering images across clothing categories due to the large spatial non-alignment of the semantics between the source and target domains; and (3) they usually require additional information to guide the image generation, e.g., the attributes of the clothing \cite{liu2019toward, liu2019collocating}, collected data of user preferences \cite{yu2019personalized}, or reference masks \cite{outfitgan,coutfitgan}. For the first issue, in practice, it is more desirable to provide multiple, diverse synthesized results rather than only one result to fashion designers: this will allow them to select their favorite ones, which can be further elaborated and used in their target design items.  For the second issue, the existing I2I translation methods cannot be simply adapted to perform the generation of compatible fashion items, as there is no apparent spatial alignment between them. In response to the third issue, a more flexible generation model without prior conditions should be developed, as users may have difficulties acquiring the additional information as model input in real-world applications. In particular, a model that has to synthesize multiple images at one time can be seen as a kind of multimodal I2I translation method \cite{huang2018multimodal,lee2018diverse,lee2020drit++,choi2020starganv2,mao2022continuous} residing in fashion domains. Generally, the aim of current multimodal I2I translation methods is to learn a distribution that explores multiple possible images in the target domain based on a given image from the source domain as input. Recently, a series of studies \cite{huang2018multimodal,lee2018diverse,lee2020drit++,choi2020starganv2,mao2022continuous} have been explored for multimodal I2I translation methods. Unfortunately, when these methods are directly applied to visually-collocated and diverse clothing synthesis, they often encounter their Waterloos. The rationale behind this may be ascribed to the following facts. Unlike the problem setting of these general methods, two crucial issues are faced in our task: (1) The translation of `upper $\rightleftharpoons$ lower' clothing has no explicit spatial alignment, in contrast to other general I2I translations, e.g., `summer $\rightleftharpoons$ winter' translation there actually is a spatial alignment in pixel level; and (2) since general multimodal I2I translation methods usually have no ground truths in the target domain, they can only learn the mapping from a source domain to a target domain in an unsupervised manner. For the first issue, these multimodal I2I methods find it hard to resolve the problem of semantic non-alignment due to the fact that they usually adopt an encoder--decoder (or U-Net) architecture, making it impossible for them to be directly applied to our task. For the second issue, our task constitutes many possible ground truths naturally, i.e., matching pairs of upper and lower clothing, which should be fully used to guide the clothing synthesis process. To properly address the above issues in our task, we propose a new framework, named \textit{BC-GAN}, to synthesize multiple compatible clothing simultaneously. Our model is built upon a pre-trained model, which helps us encode images into embeddings lying in the pre-trained latent space. The embeddings are fed into a pre-trained model to produce the targeted images. Our model is able to address the issues of spatial semantic non-alignment by taking advantage of the formulated embeddings and the generation ability of the pre-trained model. We also develop a new compatibility discriminator with a supervision ability to further improve the fashion compatibility of the synthetic results in comparison to that of previous studies \cite{liu2019toward,liu2019collocating,yu2019personalized}. A large-scale dataset containing 31,631 outfits was constructed to train and evaluate the proposed model. Extensive experiments were conducted to validate its effectiveness with respect to various evaluation metrics, in comparison to current state-of-the-art methods. Fig. \ref{cover} illustrates certain samples synthesized by our BC-GAN. For each given fashion item, BC-GAN can produce a set of diversified items. In particular, visually-collocated pairs of fashion items have been composed with a given item in both `upper $\rightarrow$ lower' and `lower $\rightarrow$ upper' directions.
	
	To sum up, the main contributions of this research are:
	
	\begin{enumerate}
		\item To the best of our knowledge, we are the first to develop a generation framework, named BC-GAN, to synthesize multiple collocated fashion items at one time. By leveraging the power of a pre-trained model, BC-GAN can accurately learn the mapping between the upper and lower clothing domains, overcoming the spatial non-alignment of the semantics during translation. 
		\item We develop a contrastive learning-based compatibility discriminator to further improve the compatibility between the given and the synthesized fashion items. Specifically, their compatibility relations are modeled from a contrastive learning perspective, in order to further enrich the supervision information for the developed discriminator. 
		\item We construct a large-scale fashion dataset \textit{DiverseOutfits}, which consists of a large number of compatible and diverse outfits. Extensive experiments validate that our proposed method can synthesize multiple compatible clothing based on a given item of clothing. Qualitative and quantitative results confirm that BC-GAN generates better results than the state-of-the-art methods in terms of diversity, visual authenticity, and fashion compatibility.
	\end{enumerate}
	
	The remainder of this paper is structured as follows. Section \ref{related_work} reviews the related work, followed by the introduction of the proposed BC-GAN in Section \ref{method}. Section \ref{experiments} details the construction of the dataset and presents the experimental results. Finally, some conclusions are drawn from our research in Section \ref{conclusion}.
	
	\section{Related Work}
	\label{related_work}
	
	Our work is most related to three streams of research, i.e., fashion compatibility learning, multimodal I2I translation, and GAN inversion. In this section, we first briefly review the research related to these three realms. Then we highlight the features of this work in comparison to the existing ones.
	
	\textbf{Fashion Compatibility Learning}. Fashion compatibility learning aims to address the collocation issues of a composed outfit. In general, current studies can be categorized into two sectors: discriminative models and generation models. For the first sector, many studies \cite{mcauley2015image,veit2015learning,vasileva2018learning,han2017learning,cui2019dressing,li2020hierarchical} focused on determining whether an outfit is compatible or not by exploring deep learning models. McAuley \textit{et al.} \cite{mcauley2015image} first used a pre-trained convolutional neural network (CNN) \cite{zhao2020scgan,xie2021dual,peng2022robust} to extract the features of fashion items, and then calculated their in-between distance to evaluate their compatibility. At the same time, Veit \textit{et al.} \cite{veit2015learning} adopted Siamese CNNs to formulate better style representations for fashion items. Later, in the work of \cite{vasileva2018learning}, a type-aware network, was proposed to more accurately model fashion compatibility. These methods which are dependent on metric learning \cite{chopra2005learning}, attempted to encode all fashion items into feature embeddings to facilitate the calculation of the distance between pairs of items in an outfit. On the other hand, Han \textit{et al.} \cite{han2017learning} argued that fashion compatibility can be modeled in a sequence model from the perspective of human observation. They adopted a bidirectional long short-term memory (Bi-LSTM) network \cite{bi_lstm} to improve the measuring of fashion compatibility. In recent years, graph neural networks (GNNs) \cite{cui2019dressing,li2020hierarchical} as emerging topics in computer vision, have been employed to characterize the relationships among fashion items. Apart from the aforementioned discriminative models, many investigations \cite{liu2019toward,liu2019collocating,yu2019personalized,outfitgan,coutfitgan} have paid considerable attention to synthesizing visually-collocated clothing by generation models, such as GAN \cite{goodfellow2014generative}. These methods regard the synthesis of collocated clothing as an I2I translation task. Specifically, they attempted to synthesize the images of complementary fashion items based on those of the given items. For example, Liu \textit{et al.} \cite{liu2019toward} first developed an Attribute-GAN framework to synthesize compatible fashion items based on a given item using attribute annotations with `upper $\rightarrow$ lower' and `lower $\rightarrow$ upper' directions. Subsequently, they further extended Attribute-GAN with a multi-discriminator architecture to improve its collocated clothing synthesis \cite{liu2019collocating}. In contrast, Yu \textit{et al.} \cite{yu2019personalized} developed a model that used the data associated with user preferences regarding outfits to facilitate personalized design for customers. In particular, these studies only worked on the translation between upper and lower clothing images to perform fashion collocation. Recently, Zhou \textit{et al.} \cite{outfitgan,coutfitgan} developed frameworks for synthesizing a set of complementary fashion items to compose an outfit with given items. However, all of these studies constitute unimodal frameworks that can only produce one complementary clothing image at a time.

	\textbf{Multimodal I2I Translation}. Multimodal I2I translation has the task of learning a conditional distribution that explores multiple possible images in a target domain, given an input image in a source domain. In order to produce diverse results, multimodal methods usually take randomly sampled noises as additional inputs. In particular, Huang \textit{et al.} \cite{huang2018multimodal} proposed a multimodal I2I translation framework, called MUNIT, to disentangle images into style and content codes in a high-level feature space. During the same period, Lee \textit{et al.} \cite{lee2018diverse} proposed a disentangled representation framework for I2I translation (DRIT) to synthesize diverse images based on the same input. In line with this study, an improved version, named DRIT++ \cite{lee2020drit++}, extended the original framework to multiple domains for synthesizing better results. Furthermore, Choi \textit{et al.} \cite{choi2020starganv2} proposed a framework with a star-shape translation network for synthesizing diverse results, especially for human faces or animal faces. More recently, a novel model \cite{mao2022continuous}, named SAVI2I, was proposed to achieve a continuous translation between source and target domains by using signed vectors. All of these methods were trained in an unsupervised manner due to the lack of ground truths in the investigated task.
	
	\textbf{GAN inversion}. GAN inversion \cite{zhu2020domain,richardson2021encoding,tov2021designing,xia2022gan} is a technique that inverts an image back into the latent space of a pre-trained GAN model as a latent code so that the visual attribute of a given image can be manipulated by editing the inverted latent code. Generally, inversion methods include optimizing the latent codes to minimize the loss for the given images \cite{tewari2020stylerig}, training an encoder to map the given images into the latent space  \cite{zhu2020domain,richardson2021encoding,tov2021designing}, or use a  hybrid method combining both strategies \cite{bau2019seeing}. The inverted latent codes can be edited by finding linear directions that correspond to changes in a given binary attribute, such as `young $\leftrightarrow$ old', or `male $\leftrightarrow$ female' \cite{zhu2020domain,richardson2021encoding,tov2021designing}. These mentioned methods manipulate the attribute of the given images by editing the inverted latent code with a pre-trained support vector machine (SVM)  \cite{shen2020interfacegan} and can be only performed in the same image domain. The fundamental principle behind GAN inversion is the disentanglement of the space of a pre-trained StyleGAN in the dimension of visual image attributes. Consequently, the inverted latent codes can be manipulated via a pre-trained SVM. It is worth noting that these inverted latent codes can only be edited and subsequently fed into the pre-trained generator for unimodal I2I translation to accomplish visual attribute editing.
	
	\textbf{Positioning of Our Work}. Among the studies discussed above, the closest works to ours are the models proposed by \cite{liu2019toward}, \cite{liu2019collocating}, and \cite{yu2019personalized}. Our task is expected to take an image of given clothing as input and produce diverse and compatible images of clothing, e.g., `upper $\rightarrow$ lower'. It falls in the category of multimodal I2I translation methods. Most multimodal I2I translation methods \cite{huang2018multimodal,lee2018diverse,lee2020drit++,choi2020starganv2,mao2022continuous} usually generate diverse images at one time in an unsupervised learning setting, and the semantics of inputs and outputs of these models have an explicit spatial alignment. For example, the `summer $\rightleftharpoons$ winter' translation \cite{huang2018multimodal,lee2018diverse,lee2020drit++,mao2022continuous} relies on pixel-level spatial alignment, while the human face translation \cite{lee2018diverse,lee2020drit++,choi2020starganv2,mao2022continuous} is implemented by hidden landmark alignment. Unfortunately, the translation task in this research, `upper $\rightleftharpoons$ lower' clothing, cannot be based on this due to the lack of explicit spatial semantic alignment between upper and lower clothing items. To alleviate the spatial nonalignment between the input and output in the task of diverse and collocated clothing synthesis. We adopt GAN inversion technique to encode images into vectors to ignore the effect of spatial information. It should be noted here that our framework is different from the GAN inversion from three aspects: (i) the composed components of our BC-GAN is different from those of GAN inversion, the additional pre-trained SVM \cite{shen2020interfacegan} to perform visual attribute manipulation is not needed in our framework; (ii) the motivation of our BC-GAN is different from that of GAN inversion. Our aim is to encode images into vectors to ignore the effect of spatial information but the insight of GAN inversion is the $\mathcal{W}$ space is disentangled about visual attributes; and (iii) the GAN inversion only supports unimodal I2I translation in visual attributes but our BC-GAN supports multimodal I2I translation in collocated clothing synthesis.
	In addition, current multimodal I2I translation methods can only use unsupervised strategies to train their models, as there are no ground truths in a training dataset. On the contrary, the availability of existing matching pairs of clothing items can be fully exploited by transforming them as a type of weakly supervised information in comparison with those unsupervised methods.
	
	\section{BC-GAN}
	\label{method}
	
	In this section, we first formulate our research problem formally. Afterward, to elaborate on the key elements in the design of our framework, we discuss related network architectures, including StyleGAN \cite{karras2019style} and its variants \cite{zhu2020domain,richardson2021encoding,tov2021designing}. Then, we introduce the overall framework of BC-GAN. The training objectives and the implementation details for training our model are then illustrated.
	
	\begin{figure*}[!]
		\centering
		\includegraphics[width=0.9\textwidth]{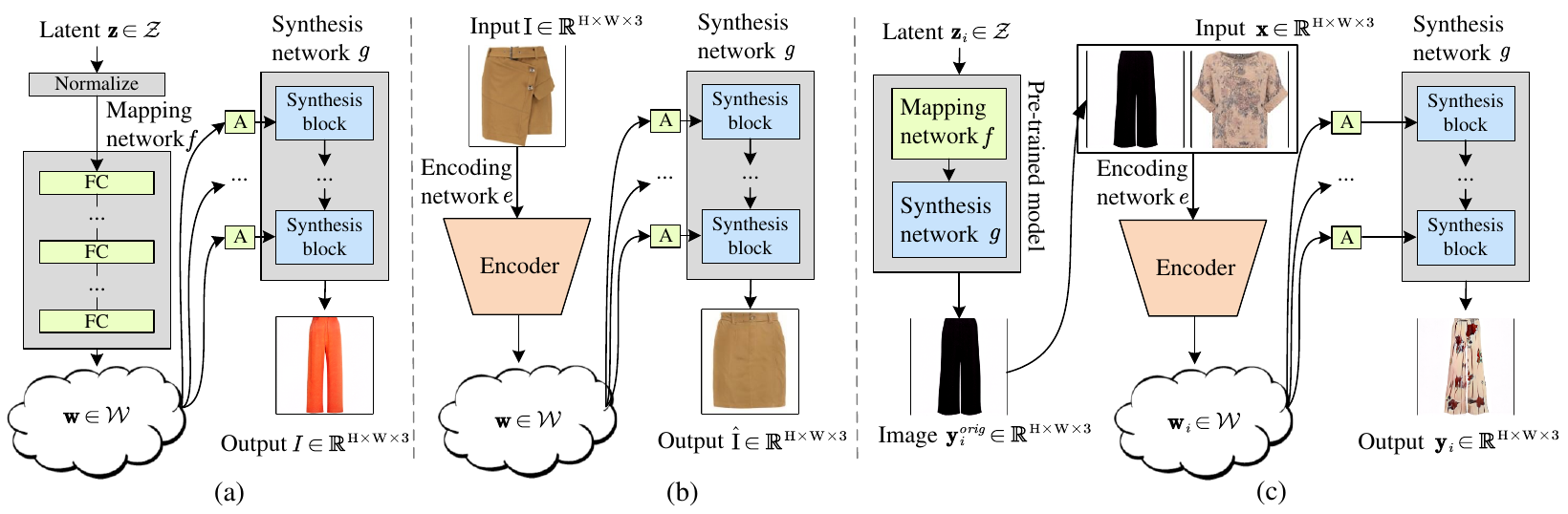}
		\caption{Comparable frameworks of the StyleGAN-based models: (a) StyleGAN \cite{karras2019style}, (b) GAN inversion for edit framework \cite{zhu2020domain}, and (c) BC-GAN (ours). Here, the pre-trained model contains mapping network $f$ and synthesis network $g$.}
		\label{cmp_methods}
		\vspace{-0.4cm}
	\end{figure*}
	
	\subsection{Problem Formulation}
	
	In general, the focus of previous collocated clothing synthesis methods \cite{liu2019toward,liu2019collocating,yu2019personalized} is leveraging additional condition information to synthesize unimodal fashion items which are expected to be compatible with given items. To achieve this, such methods usually employ an encoder--decoder (or U-Net) architecture to generate visually-collocated clothing. In contrast, recent pre-trained generative models enable us to effectively overcome the issues of semantic non-alignment and synthesize multimodal compatible fashion items. By taking advantage of these pre-trained models, our task aims at synthesizing diverse fashion items, which are compatible with a given fashion item, without using any additional information as input. As a specific multimodal I2I translation task in the fashion domain, at the current stage, we work on two translation directions, i.e., `upper $\rightarrow$ lower' and `lower $\rightarrow$ upper'. This means that given an upper (or lower) clothing item, our framework can learn a mapping capable of synthesizing multiple different lower (or upper) clothing items that are compatible with the given one. To ensure our framework can synthesize diverse results, random noise is needed to feed into our model, as widely performed in previous studies \cite{huang2018multimodal,lee2018diverse,lee2020drit++,mao2022continuous}. Formally, a mapping function in I2I translation can be represented as $F: \mathcal{X} \boldsymbol
	{\times} \mathcal{Z} \rightarrow \mathcal{Y}$, where $\mathcal{X}$, $\mathcal{Z}$ and $\mathcal{Y}$ denote a source image domain, a given noise distribution and a target image domain, respectively; and $\boldsymbol{\times}$ is the Cartesian product. Take the `upper $\rightarrow$ lower' clothing translation as an example in this research, $\mathcal{X}$ and $\mathcal{Y}$ are the upper and lower clothing domains, respectively; lying in $\mathbb{R}^{H\times W\times 3}$, where $H$ and $W$ denote the height and width of the images, respectively. $\mathcal{Z}$ lies in $\mathbb{R}^{L}$, where $L$ is the dimension of each latent code in the latent space. The translation `lower $\rightarrow$ upper' follows a similar definition. The optimization objective of our task constitutes three perspectives on the generation results, namely, diversity, visual authenticity, and fashion compatibility. Formally, given a set of $n$ latent codes $\mathbf{z}_1,\cdots,\mathbf{z}_n\in \mathcal{Z}$ and a given source clothing item $\mathbf{x}\in \mathcal{X}$, the mapping function becomes $F:\{\mathbf{x}, \mathbf{z}_i\} \mapsto \mathbf{y}_i$, where $\mathbf{z}_i$ is the $i$-th sampled noise and $\mathbf{y}_i$ is the $i$-th synthesized result. In essence, we argue that an ideal framework for diverse and compatible clothing generation should make $\mathbf{y}_i$ fit the distribution of $\mathcal{Y}$, make $\mathbf{x}$ and $\mathbf{y}_i$ compose a visually-collocated outfit, and make $\{\mathbf{y}_i\}_{i=1}^{n}$ possess a large variance in terms of a human visual perspective.
	
	\subsection{Pre-trained Models for BC-GAN}
	\label{overview}
	As mentioned, previous methods on multimodal image translation \cite{huang2018multimodal,lee2018diverse,lee2020drit++,choi2020starganv2,mao2022continuous} have difficulties addressing semantic misalignment issues due to the use of encoder–decoder architectures. To address this, we propose a new generation framework, namely, BC-GAN, which is built upon a pre-trained StyleGAN \cite{karras2019style}. For clarity, we first briefly introduce StyleGAN and its variants \cite{zhu2020domain,richardson2021encoding,tov2021designing,zhou2023fcboost}. We will then propose the generation module of our framework and further highlight the features of our work in comparison to existing studies.

	The generator architecture of StyleGAN adopts adaptive style manipulation and progressive growth to synthesize photo-realistic images. As shown in Fig. \ref{cmp_methods} (a), StyleGAN designed a generator for unconditional image generation with two key components: a mapping network $f$ and a synthesis network $g$. The mapping network $f$ constitutes a multi-layer perceptron (MLP) with eight layers. Here, $f$ is responsible for mapping a latent code $\mathbf{z}\in\mathbb{R}^{512}$, sampled from a Gaussian distribution to a style embedding $\mathbf{w}$ lying in the space $\mathcal{W}$, where $\mathcal{W}$ is a disentangled latent space. Then, the style embedding $\mathbf{w}$ is fed into the synthesis network $g$ to obtain an image $I$. By now, the rapid evolution of the StyleGAN series \cite{karras2019style,karras2020analyzing,karras2020training} has spawned many related studies \cite{zhu2020domain,richardson2021encoding,tov2021designing}, which have tried to understand and control the learned latent spaces, in order to achieve accurate attribute manipulation. As shown in Fig. \ref{cmp_methods} (b), Zhu \textit{et al.} \cite{zhu2020domain} proposed an additional encoding network $e$ to encode images beyond the pre-trained model. Their training strategy is similar to that of variational auto-encoders (VAEs) \cite{kingma2013auto} in a self-supervised manner. During the training phase, their encoders were trained with supervised loss functions, e.g., L1 loss \cite{isola2017image}, learned perceptual image patch similarity (LPIPS) \cite{zhang2018perceptual} loss, and identification (ID) loss \cite{deng2019arcface}. During the test phase, these methods aim to perform I2I translation by adopting the key idea of ``encoding first, edit later'' \cite{zhu2020domain,richardson2021encoding,tov2021designing}. Images are first encoded into a style embedding $\mathbf{w}$ using the above-mentioned encoder $e$. Then an attribute edit can be performed by moving $\mathbf{w}$ along a certain direction in the $\mathcal{W}$ space. It is worth noting that some methods \cite{richardson2021encoding,tov2021designing} use an extended space, $\mathcal{W}+$, to obtain the edited style embedding $\hat{\mathbf{w}}$, which is then fed into the pre-trained synthesis network $g$ to generate an edited image $\hat{I}$. However, they cannot be directly applied to our diverse and compatible clothing synthesis task. We attribute the underlying reason for this to the following three factors. First and foremost, the strategy for training such an encoder needs one-to-one mapping pairs, which is unachievable in our research problem, because an upper clothing item may be compatible with multiple lower clothing items according to different dressing scenarios and personal preferences. Besides, in the `upper $\rightleftharpoons$ lower' task, the edit operation performed on the style embedding cannot be implemented in the clothing translation domain due to the necessity of training an edit network in a single domain. Last but not least, our task aims at realizing multimodal I2I translation in clothing domains, while those existing frameworks only support unimodal image generation.

	To leverage the power of a pre-trained model, we propose a new paradigm for multimodal I2I translation, which is specialized for working on multimodal `upper $\rightleftharpoons$ lower' translation tasks. In particular, the generator of our framework, $G$, consists of three key components, including a mapping network $f$, an encoding network $e$, and a synthesis network $g$, as illustrated in Fig. \ref{cmp_methods} (c). Specifically, for a batch of random latent codes $\{\mathbf{z}_i\}_{i=1}^n$ sampled from a normal Gaussian distribution $\mathcal{N}(0,{\rm I})$, we first feed them into a pre-trained model to obtain diverse images lying in the target domain, which can be formulated as $f\circ g (\mathbf{z}_i) \mapsto \tilde{\mathbf{y}}_i$, where $1\leq i\leq n$. Then we concatenate this image $\tilde{\mathbf{y}}_i$ and a given image $\mathbf{x}$ from domain $\mathcal{X}$. Next, the encoder $e$ is used to encode the above-concatenated images into a style embedding $\mathbf{w}$ lying in $\mathcal{W}$. The encoded embedding $\mathbf{w}$ is fed into the pre-trained synthesis network $g$, in order to synthesize an image $\mathbf{y}_i$ lying in the domain $\mathcal{Y}$. With this design, the generation framework can effectively support multimodal I2I translation and leverage the power of the pre-trained model, as we take advantage of the generation ability of the pre-trained model.
	
	\subsection{Proposed Framework}
	
	\begin{figure*}[!t]
		\centering
		\includegraphics[width=0.9\textwidth]{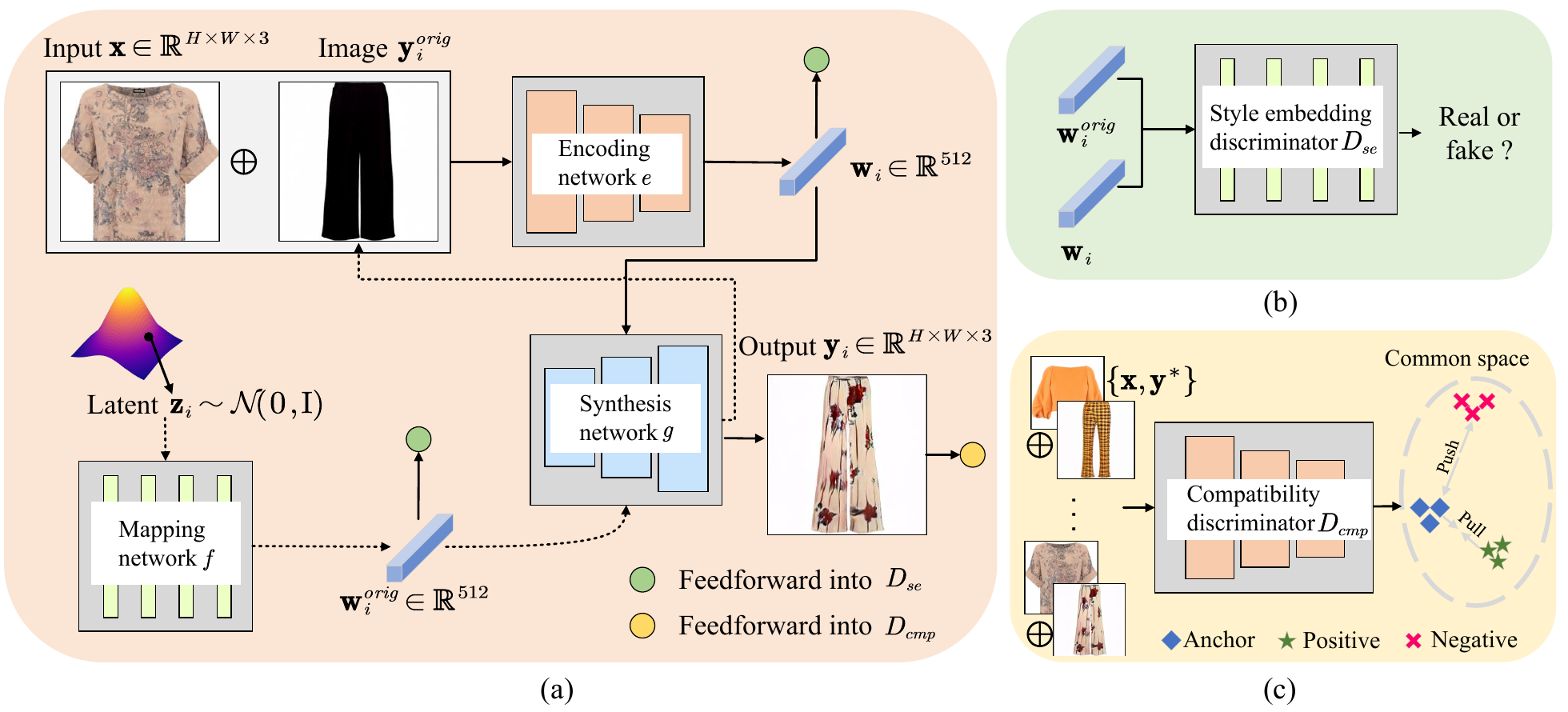}
		\caption{Key components of our proposed BC-GAN: (a) The generator $G$ which translates an input image $\mathbf{x}$ and a randomly sampled latent code $\mathbf{z}_i$ into an output image $\mathbf{y}_i$, (b) the style embedding discriminator $D_{se}$ which distinguishes between real and fake style embeddings, and (c) the compatibility discriminator $D_{cmp}$ which provides fashion compatibility supervision for synthesized clothing images in a contrastive learning perspective.}
		\label{bcgan}
		\vspace{-0.4cm}
	\end{figure*}
	
	In line with the basic generation architecture of the proposed BC-GAN, in this subsection, we introduce the detailed architecture of our BC-GAN, which is elaborated in Fig. \ref{bcgan}. The optimization process of BC-GAN will be detailed in Sections \ref{train_losses} and \ref{train_alog}. In particular, BC-GAN consists of three key modules described in the following.
	
	\textbf{Generator}. The generator $G$ shown in Fig. \ref{bcgan} (a) is tasked with translating an input image $\mathbf{x}$ and a set of sampled latent codes $\{\mathbf{z}_i\}_{i=1}^{n}$ into a set of output images $\{\mathbf{y}_i\}_{i=1}^{n}$. It contains three key components, as described in Section \ref{overview}, namely, a mapping network $f$, an encoding network $e$, and a synthesis network $g$. In particular, $f$ and $g$ were pre-trained on the target domain $\mathcal{Y}$ using the training set. The detailed encoding network of our $e$ is constructed upon that of pSp \cite{richardson2021encoding}. Formally, our generator is tasked with synthesizing a set of images $\{\mathbf{y}_i\}_{i=1}^{n}$. For each input image $\mathbf{x}$ and a given random noise $\mathbf{z}_i$, a new image $\mathbf{y}_i$ can be synthesized by the generator. The detailed feed-forward of $G(\mathbf{x}, \mathbf{y}_i)$ is described in the following three steps. First, each latent code $\mathbf{z}_i$ is fed into $f$ to obtain the original style embedding $\mathbf{w}_i^{orig}$ in the $\mathcal{W}$ space; $\mathbf{w}_i^{orig}$ is then fed into $g$ for synthesizing an original image $\mathbf{y}_i^{orig}$ lying in domain $\mathcal{Y}$. Second, the input image $\mathbf{x}$, which is concatenated with $\mathbf{y}_i^{orig}$ in channel dimension, is fed into $e$ to obtain the encoded style embedding $\mathbf{w}_i$. Finally, $\mathbf{w}_i$ is fed into $g$ for synthesizing our targeted image $\mathbf{y}_i$. Formally, for each synthesized $\mathbf{y}_i$, it can be obtained by $g(e(\mathbf{x} \mathlarger{\oplus}(f\circ g (\mathbf{z}_i))))\mapsto \mathbf{y}_i$, where $\mathlarger{\oplus}$ is a concatenation operation in channel dimension. It is worth noting that our generator also supports continuous translation by using linear interpolation between $\mathbf{w}_i^{orig}$ and $\mathbf{w}_i$.
	
	\textbf{Style embedding discriminator}. To ensure the visual authenticity of the synthesized results of BC-GAN, a real/fake discriminator needs to be developed to supervise the generator. In fact, we have two choices for achieving this goal: (i) employing a network to supervise the generated images fitting the distribution of $\mathcal{Y}$, or (ii) employing a network to supervise the encoded style embeddings fitting the distribution of $\mathcal{W}$. In particular, the second alternative which has been adopted in this research allows a model to keep the disentanglement characteristic of the encoded style embedding. Moreover, it is more efficient during model training, as the use of embeddings only needs to calculate the real/fake loss in $\mathbb{R}^{L}$ rather than using the final output images with a computational requirement in $\mathbb{R}^{H\times W \times 3}$. Here, we present a style embedding discriminator $D_{se}$ illustrated in Fig. \ref{bcgan} (b) to ensure the visual authenticity, due to the fact that the latent code $\mathbf{w}_i$ in the $\mathcal{W}$ space can favor the implementation of continuous translation in the target domain (see Section \ref{additional_study}). Our style embedding discriminator consists of an MLP with four layers. To keep the training process more stable, we also used a discriminator pool to preserve style embeddings. This discriminator is responsible for supervising the encoded style embeddings in the $\mathcal{W}$ space.
	
	\textbf{Contrastive learning-based compatibility discriminator} (Fig. \ref{bcgan} (c)). For the design of the fashion compatibility discriminator, previous studies \cite{liu2019toward,liu2019collocating,yu2019personalized} adopted a scheme by concatenating upper and lower clothing as a tuple which is fed into a discriminator directly. During this process, the compatibility discriminator only receives tuples of real pairs and synthesized pairs. The compatibility information cannot be fully exploited in this design, because the training of their compatibility discriminators has not considered the incompatibility relations which can be exploited in the training set. Based on contrastive learning \cite{hadsell2006dimensionality,schroff2015facenet,chen2020simple,xu2023spatiotemporal,xu2023pyramid,yan2023progressive,shu2022multi,xu2022x}, a compatibility discriminator can be learned by using a representation to push ``negative'' pairs further away and draw ``positive'' pairs closer in a common space. Motivated by this, we propose a new contrastive learning-based compatibility discriminator $D_{cmp}$. This discriminator is designed under the framework of Wasserstein GAN \cite{arjovsky2017wasserstein}. The discriminator can be learned from a contrastive learning perspective. In previous studies \cite{mcauley2015image,veit2015learning,vasileva2018learning,han2017learning,cui2019dressing,li2020hierarchical}, the compatible pairs of clothing items were defined as those composed by fashion experts, whereas the incompatible pairs of clothing were defined as those composed with random combinations. In our implementation, all pairs for training the compatibility discriminator can be divided into three types, $\{\mathbf{x}, \mathbf{y}^{r,c}\}$, $\{\mathbf{x}, \mathbf{y}^{r,!c}\}$, and $\{\mathbf{x}, \mathbf{y}^{f,c}\}$, where $\mathbf{x}$ is the given clothing from $\mathcal{X}$, $\mathbf{y}^{r,c}$ is a sampled item of clothing from $\mathcal{Y}$ which is compatible with $\mathbf{x}$, $\mathbf{y}^{r,!c}$ is a sampled clothing item which is not compatible with $\mathbf{x}$, and $\mathbf{y}^{f,c}$ is an item of clothing synthesized by BC-GAN, which is expected to be compatible with $\mathbf{x}$. In particular, the former two sampled items of clothing in the target domain are all selected from the training set. The training of the compatibility discriminator should ensure that the discriminator can distinguish incompatible and compatible pairs as much as possible. In contrast, all pairs of items for training our generator can be divided into three types, $\{\mathbf{x}, \mathbf{y}^{r,c}\}$, $\{\mathbf{x}, \mathbf{y}^{f,c}\}$, and $\{\mathbf{x}, \mathbf{y}^{f,!c}\}$, where $\mathbf{y}^{f,!c}$  is an item synthesized by our framework, which is not expected to be compatible with $\mathbf{x}$. Each composed pair is fed into the compatibility discriminator to obtain an embedding in a common space. For training the compatibility discriminator, we expect that the embedding of $\{\mathbf{x}, \mathbf{y}^{r,c}\}$ in the space is far from those of $\{\mathbf{x}, \mathbf{y}^{r,!c}\}$ and $\{\mathbf{x}, \mathbf{y}^{f,c}\}$. For training the generator with the compatibility loss provided by the aforementioned discriminator, the embedding of $\{\mathbf{x}, \mathbf{y}^{f,c}\}$ in the space is expected to be close to that of $\{\mathbf{x}, \mathbf{y}^{r,c}\}$, and to be far from that of $\{\mathbf{x}, \mathbf{y}^{f,!c}\}$.

	\subsection{Training Objectives}
	\label{train_losses}
	Given a batch of random latent codes $\{\mathbf{z}_i\}_{i=1}^n$ sampled from $\mathcal{Z}$ and an image $\mathbf{x} \in \mathcal{X}$, we train our framework using the following objectives.
	
	\textbf{Real/fake objective}. During training, we sample a batch of latent codes $\mathbf{z}_1, \cdots, \mathbf{z}_i, \cdots, \mathbf{z}_n \in \mathcal{Z}$ randomly and are given an image from $\mathbf{x} \in \mathcal{X}$, and generate two style embeddings: $\mathbf{w}_{i}^{orig}=f(\mathbf{z}_i)$ and $\mathbf{w}_{i}=e(\textbf{x}\oplus f\circ g(\mathbf{z}_i))$. The style embedding discriminator $D_{se}$ takes these style embeddings $\mathbf{w}_{i}^{orig}$ and $\mathbf{w}_{i}$ as inputs and learns to distinguish the original style embeddings and the encoded style embeddings with the following loss:
	\begin{equation}
		\label{dis_loss}
		\begin{aligned}
			\mathcal{L}_{dis}=~&\mathbb{E}_{\mathbf{z}_i\sim \mathcal{Z}, \mathbf{x}\sim \mathcal{X}}[\log D_{se}(f(\mathbf{z}_i)) + \\
			&\log (1-D_{se}(e(\textbf{x}\oplus f\circ g(\mathbf{z}_i))))],
		\end{aligned}
	\end{equation}
	where two terms on the right hand are responsible for $\mathbf{w}_{i}^{orig}$ and $\mathbf{w}_{i}$, respectively. The encoder $e$ takes an image $\mathbf{x}$ and a random synthesized image $f\circ g(\mathbf{z}_i)$ as inputs and learns to generate a style embedding via an adversarial loss:
	\begin{equation}
		\label{adv_loss}
		\mathcal{L}_{adv}=\mathbb{E}_{\mathbf{z}_i\sim \mathcal{Z}, \mathbf{x}\sim \mathcal{X}}[\log (D_{se}(e(\textbf{x}\oplus f\circ g(\mathbf{z}_i))))].
	\end{equation}
	With these training losses, our generator can produce $\mathbf{w}_{i}$ lying in $\mathcal{W}$ space. This real/fake objective is designed to guide our generator to synthesize photo-realistic images.
	
	\textbf{Fashion compatibility objective}. To guarantee the compatibility of a synthesized clothing item $\mathbf{y}_i$ with the given image $\mathbf{x}$, we employ a contrastive loss for training our compatibility discriminator and our generator. For training the compatibility discriminator $D_{cmp}$, we have the following training objective:
	\begin{equation}
		\label{cmp_dis_loss}
		\begin{aligned}
			\mathcal{L}_{cmp\_dis}&=\\
			&[\mathbb{E}_{\mathbf{z}_i\sim \mathcal{Z}, \mathbf{x}\sim \mathcal{X}}(D_{cmp}(\{\mathbf{x}, \mathbf{y}^{r,c}\}))\\
			-&\mathbb{E}_{\mathbf{z}_i\sim \mathcal{Z}, \mathbf{x}\sim \mathcal{X}}(D_{cmp}(\{\mathbf{x}, \mathbf{y}^{r,!c}\}))]  \\
			+&[\mathbb{E}_{\mathbf{z}_i\sim \mathcal{Z}, \mathbf{x}\sim \mathcal{X}}(D_{cmp}(\{\mathbf{x}, \mathbf{y}^{r,c}\}))\\
			-&\mathbb{E}_{\mathbf{z}_i\sim \mathcal{Z}, \mathbf{x}\sim \mathcal{X}}(D_{cmp}(\{\mathbf{x}, \mathbf{y}^{f,c}\}))],
		\end{aligned}
	\end{equation}
	where $D_{cmp}$ is functional on mapping pairs of fashion items into a common space, encoding composed outfits into embeddings, and it pushes $\{\mathbf{x}, \mathbf{y}^{r,c}\}$ far from both $\{\mathbf{x}, \mathbf{y}^{r,!c}\}$ and $\{\mathbf{x}, \mathbf{y}^{f,c}\}$ in the space. For training the generator $G$, the compatibility loss is defined as follows:
	\begin{equation}
		\label{cmp_loss}
		\begin{aligned}
			\mathcal{L}_{cmp}&=\\
			&[\mathbb{E}_{\mathbf{z}_i\sim \mathcal{Z}, \mathbf{x}\sim \mathcal{X}}(D_{cmp}(\{\mathbf{x}, \mathbf{y}^{f,c}\}))\\
			-&\mathbb{E}_{\mathbf{z}_i\sim \mathcal{Z}, \mathbf{x}\sim \mathcal{X}}(D_{cmp}(\{\mathbf{x}, \mathbf{y}^{r,c}\}))]  \\
			+&[\mathbb{E}_{\mathbf{z}_i\sim \mathcal{Z}, \mathbf{x}\sim \mathcal{X}}(D_{cmp}(\{\mathbf{x}, \mathbf{y}^{f,c}\}))\\
			-&\mathbb{E}_{\mathbf{z}_i\sim \mathcal{Z}, \mathbf{x}\sim \mathcal{X}}(D_{cmp}(\{\mathbf{x}, \mathbf{y}^{f,!c}\}))],
		\end{aligned}
	\end{equation}
	where $D_{cmp}$ draws $\{\mathbf{x}, \mathbf{y}^{f,c}\}$ close to $\{\mathbf{x}, \mathbf{y}^{r,c}\}$ and pushes $\{\mathbf{x}, \mathbf{y}^{f,c}\}$ away from $\{\mathbf{x}, \mathbf{y}^{f,!c}\}$.
	
	\textbf{Diversity objective.} To further enable the generator $G$ to produce diverse clothing images $\{\mathbf{y}_{i}\}_{i=1}^{n}$ belonging to $\mathcal{Y}$, we explicitly regularize $G$ with the following diversity loss, in line with \cite{choi2020starganv2}:
	\begin{equation}
		\label{div_loss}
		\mathcal{L}_{div}=-\mathbb{E}_{\mathbf{y}_i \neq \mathbf{y}_j}[d(\mathbf{y}_i, \mathbf{y}_j)],
	\end{equation}
	where $d(\cdot,\cdot)$ is a distance function. In our implementation, we used the LPIPS \cite{zhang2018perceptual} as our distance metric, which can guarantee convergence during the optimization process. In fact, we have also conducted experiments with the L1/L2 distance, as alternatives to LPIPS. However, according to our experiments, it is hard for the alternative models to converge. The underlying reason behind this phenomenon may be ascribed to the fact that the last layer of the pre-trained model has no activation function and the output of the last layer resides in a large range.
	
	\textbf{Full objective}. For training the generator of our BC-GAN, the full objective functions can be summarized as follows:
	\begin{equation}
		\label{total_loss}
		\mathcal{L}_{total}=\mathcal{L}_{adv}+\lambda_1 \mathcal{L}_{div} + \lambda_2 \mathcal{L}_{cmp},
	\end{equation}
	where $\mathcal{L}_{adv}$ is a GAN loss, which is responsible for the visual authenticity of the synthesized results; $\mathcal{L}_{div}$ is a diversity loss, which is responsible for synthesizing diverse results; and $\mathcal{L}_{cmp}$ is a compatibility loss, which is responsible for the fashion compatibility of the synthesized results. Moreover, $\lambda_1$ and $\lambda_2$ are parameters to balance the relative contributions of the loss terms.

	\begin{algorithm}[htbp]
		\caption{Adversarial training algorithm for BC-GAN}
		\small
		\label{alg}
		\SetAlgoLined
		\KwIn{Compatible and diverse outfit pairs $\{\mathbf{x}, \mathbf{y}\} \in \mathcal{X} \boldsymbol{\times} \mathcal{Y}$.}
		\KwOut{Generator $G$ of BC-GAN.}
		
		Pre-train StyleGAN on target domain $\mathcal{Y}$ of our training set; \textcolor[rgb]{0.5,0.5,0.5}{\# To obtain $f$ and $g$}
		
		Initialize the parameters of $e$, $D$ and $D_{cmp}$, copy the parameters of $f$ and $g$ from the pre-trained StyleGAN and freeze them;
		
		\For{$iter\leftarrow 1$ \KwTo $N_{iter}$}{
			Sample a batch of $\{\mathbf{x}, \mathbf{y}\}$ from training set;
			
			Sample a batch of random latent codes $\{\mathbf{z}_i\}_{i=1}^n$ from $\mathcal{N}(0, \mathbf{{\rm I}})$; 
			
			\textcolor[rgb]{0.5,0.5,0.5}{\# To train the style embedding discriminator $D$}
			
			%
			%
			%
			
			$\mathcal{L}_{dis}=\mathbb{E}_{\mathbf{z}_i\sim \mathcal{Z}, \mathbf{x}\sim \mathcal{X}}[\log D_{se}(f(\mathbf{z}_i)) + \log (1-D_{se}(e(\textbf{x}\oplus f\circ g(\mathbf{z}_i))))]$; \textcolor[rgb]{0.5,0.5,0.5}{ \# See Eq. (\ref{dis_loss})}
			
			update parameters $\theta_{D_{se}}$ of $D_{se}$ with
			$\theta_{D_{se}} \leftarrow \theta_{D_{se}} - \eta \bigtriangledown_{\theta_{D_{se}}}\mathcal{L}_{dis}$; \textcolor[rgb]{0.5,0.5,0.5}{\# $\eta$ is a learning rate}
			
			\textcolor[rgb]{0.5,0.5,0.5}{\# To train the compatibility discriminator $D_{cmp}$}
			
			$\mathbf{y}_i\leftarrow e\circ g(\mathbf{x}\oplus f \circ g(\mathbf{z}_i))$, where $i\in \{1,\cdots,n\}$;
			
			Assign sample $\mathbf{x}$ with samples $\mathbf{y}_i$ ($i\in\{1,\cdots, n\}$), $\mathbf{y}$ into $\{\mathbf{x}, \mathbf{y}^{r,c}\}$, $\{\mathbf{x}, \mathbf{y}^{f,c}\}$, and $\{\mathbf{x}, \mathbf{y}^{r,!c}\}$ pairs;
			
			$\mathcal{L}_{cmp\_dis}\leftarrow [\mathbb{E}_{\mathbf{z}_i\sim \mathcal{Z}, \mathbf{x}\sim \mathcal{X}}(D_{cmp}(\{\mathbf{x}, \mathbf{y}^{r,c}\}))-
			\mathbb{E}_{\mathbf{z}_i\sim \mathcal{Z}, \mathbf{x}\sim \mathcal{X}}(D_{cmp}(\{\mathbf{x}, \mathbf{y}^{r,!c}\}))] 
			+[\mathbb{E}_{\mathbf{z}_i\sim \mathcal{Z}, \mathbf{x}\sim \mathcal{X}}(D_{cmp}(\{\mathbf{x}, \mathbf{y}^{r,c}\}))-
			\mathbb{E}_{\mathbf{z}_i\sim \mathcal{Z}, \mathbf{x}\sim \mathcal{X}}(D_{cmp}(\{\mathbf{x}, \mathbf{y}^{f,c}\}))]$; \textcolor[rgb]{0.5,0.5,0.5}{\# See Eq. (\ref{cmp_dis_loss})}
			
			update parameters $\theta_{D_{cmp}}$ of $D_{cmp}$ with
			$\theta_{D_{cmp}} \leftarrow \theta_{D_{cmp}} - \eta \bigtriangledown_{\theta_{D_{cmp}}}\mathcal{L}_{cmp\_dis}$;
			
			\textcolor[rgb]{0.5,0.5,0.5}{\# To train the generator $G$}
			
			$\mathbf{w}_i^{orig}\leftarrow f(\mathbf{z}_i)$, where $i\in \{1,\cdots,n\}$;
			
			$\mathbf{w}_i\leftarrow e(\mathbf{x}\oplus g(\mathbf{w}_i^{orig}))$, where $i\in \{1,\cdots,n\}$;
			
			$\mathbf{y}_i\leftarrow e\circ g(\mathbf{x}\oplus g(\mathbf{w}_i))$, where $i\in \{1,\cdots,n\}$;
			
			Assign sample $\mathbf{x}$ with samples $\mathbf{y}_i$ ($i\in\{1,\cdots, n\}$), $\mathbf{y}$ into $\{\mathbf{x}, \mathbf{y}^{r,c}\}$, $\{\mathbf{x}, \mathbf{y}^{f,c}\}$, and $\{\mathbf{x}, \mathbf{y}^{f,!c}\}$ pairs;
			
			$\mathcal{L}_{adv}\leftarrow \mathbb{E}_{\mathbf{z}_i\sim \mathcal{Z}, \mathbf{x}\sim \mathcal{X}}[\log (D_{se}(\mathbf{y}_i)]$; \textcolor[rgb]{0.5,0.5,0.5}{\# See Eq. (\ref{adv_loss})}
			
			$\mathcal{L}_{cmp}\leftarrow [\mathbb{E}_{\mathbf{z}_i\sim \mathcal{Z}, \mathbf{x}\sim \mathcal{X}}(D_{cmp}(\{\mathbf{x}, \mathbf{y}^{f,c}\}))-
			\mathbb{E}_{\mathbf{z}_i\sim \mathcal{Z}, \mathbf{x}\sim \mathcal{X}}(D_{cmp}(\{\mathbf{x}, \mathbf{y}^{r,c}\}))] 
			+[\mathbb{E}_{\mathbf{z}_i\sim \mathcal{Z}, \mathbf{x}\sim \mathcal{X}}(D_{cmp}(\{\mathbf{x}, \mathbf{y}^{f,c}\})) -
			\mathbb{E}_{\mathbf{z}_i\sim \mathcal{Z}, \mathbf{x}\sim \mathcal{X}}(D_{cmp}(\{\mathbf{x}, \mathbf{y}^{f,!c}\}))]$; \textcolor[rgb]{0.5,0.5,0.5}{\# See Eq. (\ref{cmp_loss})}
			
			$\mathcal{L}_{div}\leftarrow -\mathbb{E}_{\mathbf{y}_i \neq \mathbf{y}_j}[d(\mathbf{y}_i, \mathbf{y}_j)]$; \textcolor[rgb]{0.5,0.5,0.5}{\# See Eq. (\ref{div_loss})}
			
			$\mathcal{L}_{total}\leftarrow \mathcal{L}_{adv}+\lambda_1 \mathcal{L}_{div} + \lambda_2 \mathcal{L}_{cmp}$; \textcolor[rgb]{0.5,0.5,0.5}{\# See Eq. (\ref{total_loss})}
			
			update parameters $\theta_{G}$ of $G$ with 
			$\theta_{G} \leftarrow \theta_{G} - \eta \bigtriangledown_{\theta_{G}}\mathcal{L}_{total}$;
		}
		\Return{$G^{*}$;} \textcolor[rgb]{0.5,0.5,0.5}{\# Optimized $G$ (includes $f$, $e$ and $g$)}
	\end{algorithm}
	
	\subsection{The Adversarial Training Process}
	\label{train_alog}
	In this subsection, we present the design of an adversarial training scheme which is used to optimize the generator $G$ of BC-GAN. For clarity, the entire training process of BC-GAN is summarized in Algorithm \ref{alg}. In the beginning, we pre-trained a StyleGAN on the target domain using our training set (shown in line 1). Then the parameters of $e$, $D_{se}$ and $D_{cmp}$ are initialized, and the parameters of $f$ and $g$ are copied from the above pre-trained StyleGAN and frozen (shown in line 2). The subsequent training process is carried out by applying a gradient descent step on $D_{se}$, $D_{cmp}$ and $G$ in alternating steps, and using the gradient descent method to update the parameters $\theta_{D_{se}}$, $\theta_{D_{cmp}}$ and $\theta_{G}$ of $D_{se}$, $D_{cmp}$ and $G$ with a learning rate $\eta$, respectively (shown in lines 3--24). In particular, given a batch of pairs of visually-collocated items of clothing $\{\mathbf{x}, \mathbf{y}\}$ and a batch of sampled latent codes $\{\mathbf{z}\}_{i=1}^{n}$ from $\mathcal{N}(0, \mathbf{{\rm I}})$ (shown in lines 4--5), in the following training steps, we first train our style embedding discriminator $D_{se}$ (shown in lines 6--8). The style embeddings $\{\mathbf{w}_i^{orig}\}_{i=1}^{n}$ from $f$ and $\{\mathbf{w}_i\}_{i=1}^{n}$ from $e$ are generated and then fed into $D_{se}$ to update its parameters (shown in lines 7--8). Subsequently, the compatibility discriminator $D_{cmp}$ is trained from a contrastive learning perspective (shown in lines 9--13). Different pairs are synthesized and designated by specified labels (shown in lines 10--11). These pairs are fed into $D_{cmp}$ to update its parameters (shown in lines 12--13). At last, the adversarial loss $\mathcal{L}_{adv}$ (shown in line 19), compatibility loss $\mathcal{L}_{cmp}$ (shown in line 20), and diversity loss $\mathcal{L}_{div}$ (shown in line 21) are calculated. Moreover, our generator $G$ is trained with $\mathcal{L}_{total}$ (shown in lines 22--23). When the training process converges, our framework returns the optimized generator $G$ of BC-GAN to synthesize compatible and diverse clothing images.
	
	\section{Experiments}
	\label{experiments}
	In this section, the construction of our dataset, named DiverseOutfits, is first introduced. Then we experimentally show the superiority of our proposed method over existing methods in terms of diversity, visual authenticity, and fashion compatibility. Furthermore, we provide an ablation study to verify the effectiveness of the main modules in BC-GAN. Finally, an additional study about continuous translation on the target domain is conducted.
	
	\subsection{Dataset}
	\label{sec_dataset}
\begin{figure}[t]		
		\centering
		\includegraphics[width=0.5\textwidth]{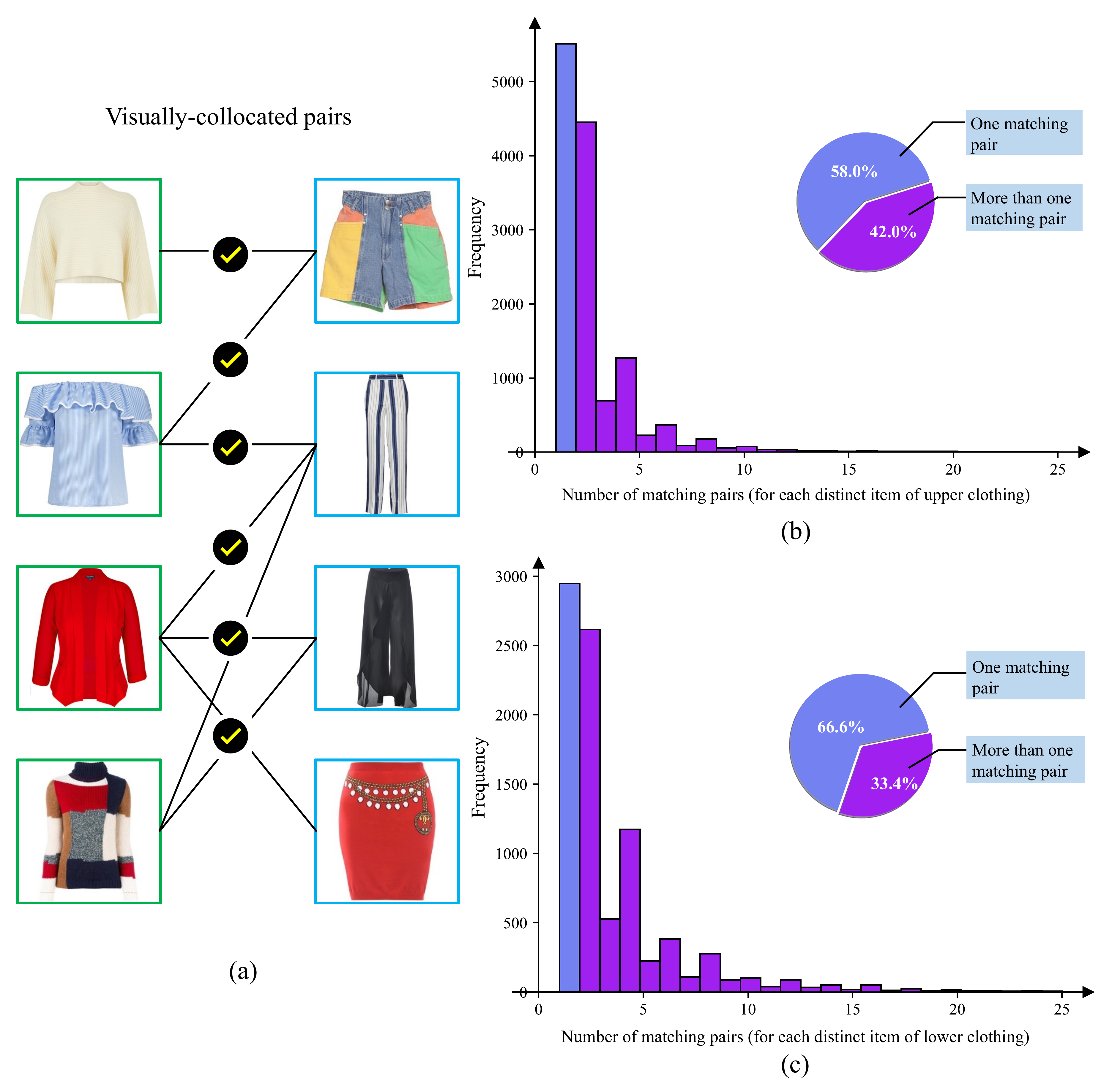}
		\caption{Statistics of our DiverseOutfits dataset: (a) an illustration of multiple possible matching outfits, (b) statistics of matching pairs (for each distinct item of upper clothing), and (c) statistics of matching pairs (for each distinct item of lower clothing).}
		\label{dataset}
		\vspace{-0.4cm}
	\end{figure}

	When carrying out collocated clothing generation, a large-scale fashion dataset which can provide valid samples for model training and test becomes essential. In fact, several fashion datasets have been constructed for different research purposes, such as UT Zappos50K \cite{yu2014fine}, the Maryland Polyvore dataset \cite{han2017learning}, and the OutfitSet dataset \cite{outfitgan}. However, most of the existing released datasets lack annotations of diverse matching pairs of fashion items. Sometimes they treat the same clothing as different entities. In order to address the issues present in the existing datasets, we have collected a comprehensive dataset, named DiverseOutfit, from a fashion matching community, Polyvore.com. This dataset includes 31,631 outfits, each of which is composed of an upper and lower clothing item. Each outfit has been meticulously curated by fashion experts to ensure compatibility.  
	In order to merge identical entities, for each category of fashion items, LPIPS was employed to gauge the visual similarity of two images of upper (or lower) clothing. If the  similarity between these two items of clothing is beyond the set cut-off, they were classified as belonging to the same entity.
	As shown in Fig. \ref{dataset} (a), we can observe that each fashion item can match with multiple complementary fashion items. To further verify our observations, we conducted statistics on the distributions in terms of an upper (or lower) clothing item matching how many lower (or upper) clothing. The statistics illustrated in Figs. \ref{dataset} (b) and (c) show that 58.0\% (or 66.6\%) of upper (or lower) clothing items have one matching fashion item, while 42.0\% (or 33.4\%) of upper (or lower) clothing have more than one matching fashion item, indicating the diverse compatible rules hidden in the DiverseOutfits dataset. We then partitioned these outfits into two subsets to form a training set and a test set. In fact, as a multimodal I2I translation method, our task has two different settings: the `upper $\rightarrow$ lower' and `lower $\rightarrow$ upper' directions. For the `upper $\rightarrow$ lower' setting, we split the outfit composition pairs into training and test sets randomly. There is no overlap for the upper clothing between the training and test sets. The training and test sets have a ratio of $4 : 1$, according to the number of matching pairs. For the `upper $\rightarrow$ lower', the training set has 25,305 matching pairs, and the test set has 6,326 matching pairs. Similarly, we constructed the dataset for the `lower $\rightarrow$ upper' setting, in which its training and test sets have 25,305 and 6,326 matching pairs, respectively.
	
	\subsection{Experimental Setup and Parameter Settings}
	In our experiments, all images were resized to 256$\times$256 and $L$ was set to $512$. During the training phase, the batch size of our BC-GAN was set to four, and the number of training iterations was set to 50,000. All experiments were performed on a single A6000 graphics card, and the implementation was carried out in PyTorch \cite{paszke2017automatic}. BC-GAN was trained with an Adam \cite{kingma2014adam} optimizer with $\beta_1=0$ and $\beta_2=0.99$, and the learning rate $\eta$ was set to $2\times 10^{-4}$. In order to make the adversarial training more stable, the style embedding discriminator pool was set with a size of 100 in our implementation. The coefficients of Eq. (\ref{total_loss}) were empirically set with $\lambda_1=1$ and $\lambda_2=3$.
	
	\subsection{Evaluation Metrics}
	\label{eva_metric}
	The generated clothing images were evaluated from three important aspects: diversity, visual authenticity, and fashion compatibility. For clarity, the evaluation metrics used are described as follows:
	
	\subsubsection{Diversity Measurement} A diversity measurement is used to measure the diversity of the synthesized images $\{\mathbf{y}_i\}_{i=1}^{n}$ lying in the target domain $\mathcal{Y}$ for the same input image $\mathbf{x}$ from the source domain $\mathcal{X}$. Following previous studies \cite{huang2018multimodal,choi2020starganv2}, we use LPIPS as our evaluation metric. It calculates the mean distance of each possible combination of synthesized images as follows:
	\begin{equation}
		{\rm LPIPS}=\mathbb{E}_{\mathbf{x}\sim \mathcal{X}}[\sum_{i=1}^{N}\sum_{j=i+1}^{N}\frac{2\times d(\mathbf{y}_i, \mathbf{y}_j)}{N\times (N-1)}],
	\end{equation}
	where $N$ is the number of synthesized images for each input image, and we set $N=10$. The distance function $d(\cdot,\cdot)$ was implemented by LPIPS with a pre-trained AlexNet\footnote{\url{https://github.com/richzhang/PerceptualSimilarity}} to evaluate the performance in terms of diversity. Here, a higher score indicates better diversity.
	\subsubsection{Visual Authenticity Measurement} Previous studies \cite{yu2019personalized,choi2020starganv2,mao2022continuous,outfitgan,coutfitgan} have suggested that the Fr\'{e}chet inception distance \cite{heusel2017gans} (FID) can be used to estimate the visual authenticity of synthesized images in a high-level feature space mapped by a pre-trained model. In particular, FID measures the similarity of the statistics of the features of synthesized and real images. We used the implementation\footnote{\url{https://github.com/mseitzer/pytorch-fid}} of FID to measure the visual authenticity of synthesized images. A lower FID value means better visual authenticity.
	
	\subsubsection{Fashion Compatibility Measurement} A fashion compatibility measurement is used to gauge the visually-collocated degree of a given clothing image and synthesized clothing image. Following Zhou \textit{et al.} \cite{coutfitgan}, we adopt fashion fill-in-the-blank best times (F${\rm ^2}$BT) as a metric to estimate the fashion compatibility of synthesized fashion items with a given item. To make this paper self-contained, a brief introduction of F${\rm ^2}$BT is given here. F${\rm ^2}$BT adopts a pre-trained fashion compatibility predictor $\varphi$ in an open-source toolbox MMFashion\footnote{\url{https://github.com/open-mmlab/mmfashion}}. It should be noted that the dataset on which $\varphi$ was trained, Maryland Polyvore dataset \cite{han2017learning}, is different from that our experiments. $\varphi$ plays a role in comparing two composed outfits as well as predicting which of them is better. We count the number of times that each method beats all other methods, which is under the concept of `one-vs-rest'. Formally, it is given as follows:
	\begin{equation}
		{\rm F^2BT}(m_i)=\mathbb{E}_{\mathbf{o}_k}[\frac{\prod_{j=1,j\neq i}^{N_{m}} \zdlmathds{1}(\varphi(\mathbf{o}_{i,k})>\varphi(\mathbf{o}_{j,k}))}{N_{m}}],
	\end{equation}
	where $\zdlmathds{1}(\cdot)$ is a conditional function; for the $i$-th method in the compared set, we count the average number of times that this method beats the other methods in terms of fashion compatibility; $\mathbf{o}_{i,k}$ is the $k$-th outfit synthesized by the $i$-th method $m_i$, and $\mathbf{o}_{j,k}$ has a similar definition. It should be noted that all values are normalized by the size of the set of compared methods. Here, a higher value means better fashion compatibility.

	\subsection{Performance Comparison}
	
	\begin{figure*}[!htbp]
		\centering
		\includegraphics[width=0.96\textwidth]{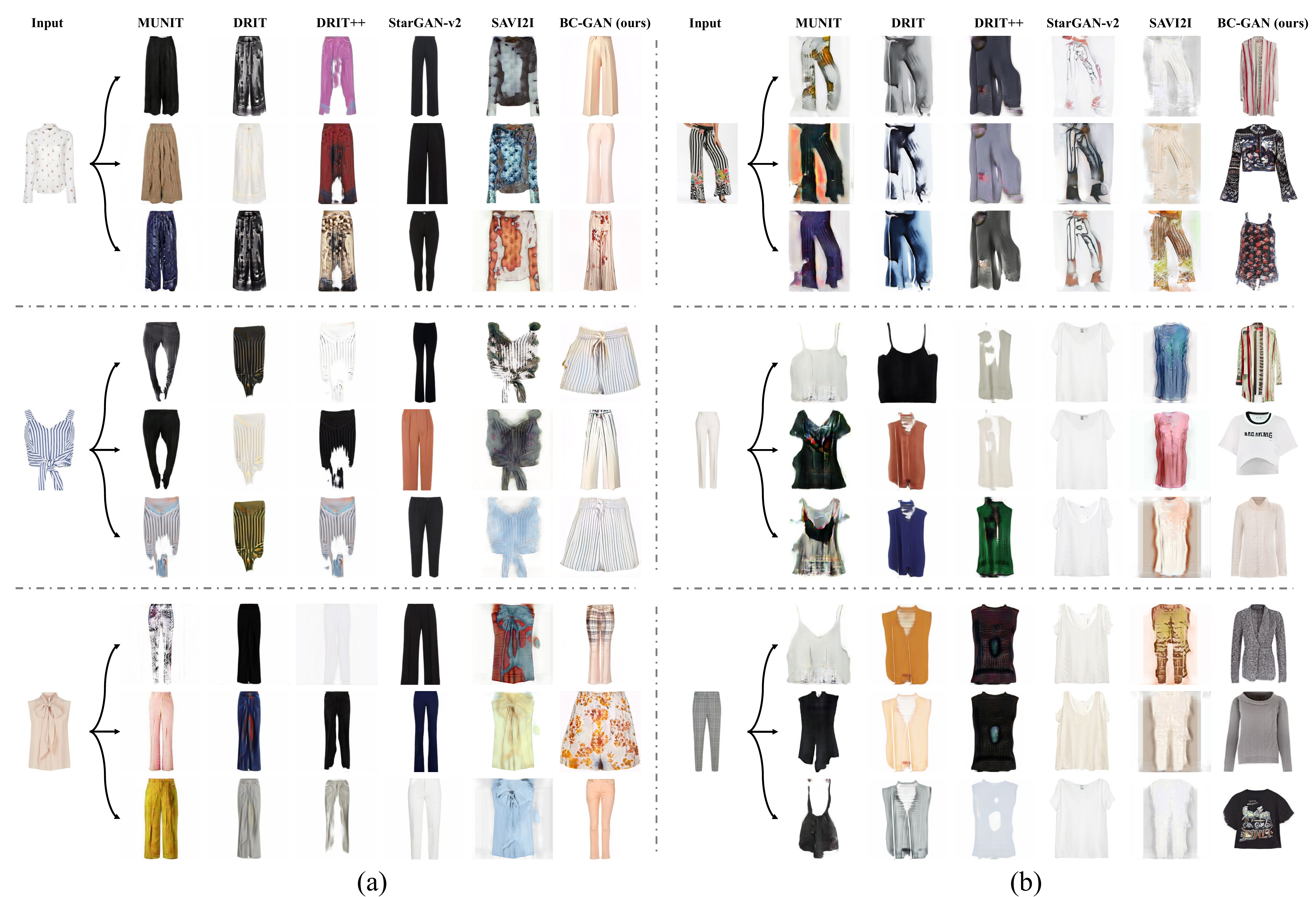}
		\caption{Comparisons between our BC-GAN and other multimodal I2I translation methods, which are MUNIT \cite{huang2018multimodal}, DRIT \cite{lee2018diverse}, DRIT++ \cite{lee2020drit++}, StarGAN-v2 \cite{choi2020starganv2}, and SAVI2I \cite{mao2022continuous}, in terms of (a) translation on upper $\rightarrow$ lower setting, and (b) translation on lower $\rightarrow$ upper setting.}
		\label{cmp_samples}
		\vspace{-0.4cm}
	\end{figure*}
	
	\subsubsection{Compared Methods}
	Since there are few other models that can carry out the task considered here, our proposed BC-GAN is evaluated against five state-of-the-art methods, which are closely related to our method. The five compared baseline methods include MUNIT \cite{huang2018multimodal}, DRIT \cite{lee2018diverse}, DRIT++ \cite{lee2020drit++}, StarGAN-v2 \cite{choi2020starganv2}, and SAVI2I \cite{mao2022continuous}. We give a brief introduction to these baselines as follows:
	
	\textbf{MUNIT} \cite{huang2018multimodal} attempts to disentangle images into content codes and style codes in a high-level feature space with two different encoders. To synthesize diverse images, it combines content codes and style codes encoded from different images. During the model training, the reconstruction loss is a key component to learn conditional mapping.
	
	\textbf{DRIT} \cite{lee2018diverse} proposes an approach based on disentangled representation for producing diverse outputs in I2I translation. It attempts to disentangle the information of an image into domain-invariant content space and domain-specific attribute space. 
	
	\textbf{DRIT++} \cite{lee2020drit++} is an improved version of DRIT. It extends the original framework from two domains to multiple domains for I2I translation. Moreover, a new mode-seeking regularization is proposed to improve the diversity of the synthesized results.
	
	\textbf{StarGAN-v2} \cite{choi2020starganv2} is an improved version of StarGAN \cite{choi2018stargan}. Compared with StarGAN, which only supports unimodal I2I translation, StarGAN-v2 supports multimodal translation.
	
	\textbf{SAVI2I} \cite{mao2022continuous} is the first framework supporting continuous multimodal I2I translation. It employs signed attribute vectors to control the continuous I2I translation.
	
	For a fair comparison, all model implementations were based on the original codes reported by the authors. For each translation direction in diverse collocated clothing synthesis, these baseline models were trained from scratch. For instance, in the `upper $\rightarrow$ lower' translation, the source and target domains are upper and lower clothing, respectively.
	
	\subsubsection{Comparison of Results}
	
	For comparison, the quantitative evaluation results of all compared methods are given in Tables \ref{upper2lower_stat} and \ref{lower2upper_stat}. Moreover, a visual comparison of synthesized results is given in Fig. \ref{cmp_samples}. As described in Section \ref{sec_dataset}, our problem has two translation directions: `upper $\rightarrow$ lower' and `lower $\rightarrow$ upper' settings. For the setting of `upper $\rightarrow$ lower' translation, the results of the comparison of BC-GAN and other baselines are given in Table \ref{upper2lower_stat}. It shows that our proposed BC-GAN consistently outperforms other multimodal I2I translation methods in terms of diversity (LPIPS), visual authenticity (FID), and fashion compatibility (F${\rm ^2}$BT). The results of `upper $\rightarrow$ lower' and `lower $\rightarrow$ upper' directions are analyzed independently as follows:

	\begin{table}[t]
		\centering
		\caption{Comparison of BC-GAN and baselines on upper $\rightarrow$ lower setting in terms of diversity (LPIPS), visual authenticity (FID), and fashion compatibility (F$\rm{^2}$BT) (here, for all metrics except FID, higher is better)}
		\label{upper2lower_stat}
		
		\begin{tabular}{p{2.5cm}  P{1.2cm} P{1.2cm} P{1.2cm} }
			\toprule
			\multirow{3}{*}{Method} & \multicolumn{3}{c}{Evaluation metrics on upper $\rightarrow$ lower}\\
			\cmidrule[0.5pt]{2-4}
			&LPIPS($\uparrow$)&FID($\downarrow$)  & F$\rm{^2}$BT($\uparrow$)*\\
			\hline
			MUNIT \cite{huang2018multimodal} &  0.479 &  60.016 & 16.5\%\\
			DRIT \cite{lee2018diverse} & 0.248 & 93.359 & 14.2\%\\
			DRIT++ \cite{lee2020drit++} &  0.304 & 74.033 & 14.0\%\\
			StarGAN-v2 \cite{choi2020starganv2} &   0.343 & 92.489 & 13.6\%\\
			SAVI2I \cite{mao2022continuous} &   0.397 & 123.944 & 20.0\%\\
			\hline
			BC-GAN (ours)&  \textbf{0.504} & \textbf{52.370} & \textbf{22.9\%}\\
			\bottomrule
		\end{tabular}
		
		* Note: Percentages may not sum to 100 because of rounding.
		\vspace{-0.4cm}
	\end{table}
	
	\begin{table}[t]
		\centering
		\caption{Comparison of BC-GAN and baselines on lower $\rightarrow$ upper setting in terms of diversity (LPIPS), visual authenticity (FID), and fashion compatibility (F$\rm{^2}$BT) (here, for all metrics except FID, higher is better)}
		\label{lower2upper_stat}
		\begin{tabular}{p{2.5cm} P{1.2cm} P{1.2cm} P{1.2cm} }
			\toprule
			\multirow{3}{*}{Method} & \multicolumn{3}{c}{Evaluation metrics on lower $\rightarrow$ upper}\\
			\cmidrule[0.5pt]{2-4}
			&LPIPS($\uparrow$)&FID($\downarrow$)&F$\rm{^2}$BT($\uparrow$)*\\
			\hline
			MUNIT \cite{huang2018multimodal} &  0.547 & 47.036 & 16.6\%\\
			DRIT \cite{lee2018diverse} &  0.263 & 77.858 & 16.1\%\\
			DRIT++ \cite{lee2020drit++} &  0.335 & 61.170 & 14.4\%\\
			StarGAN-v2 \cite{choi2020starganv2} &  0.386  & 42.626 & 14.7\%\\
			SAVI2I \cite{mao2022continuous} &  0.378 & 118.546 & 15.4\%\\
			\hline
			BC-GAN (ours)&  \textbf{0.605} & \textbf{38.981} & \textbf{22.8\%}\\
			\bottomrule
		\end{tabular}
		
		* Note: Percentages may not sum to 100 because of rounding.
		\vspace{-0.4cm}
	\end{table}
	
 \textbf{The `upper $\rightarrow$ lower' direction:} 1) For the metric of diversity, our method improves by approximately 0.025 in comparison with the second-best method. From the synthesized results in Fig. \ref{cmp_samples} (a), it is clear that our model produces the most diverse results, especially for shape and texture. Taking the second and third samples for the `upper $\rightarrow$ lower' setting as an example, we can observe that BC-GAN synthesizes shorts, skirts, or trousers with diverse shapes and textures. Compared with our method, the results of other baselines are not as promising as ours in terms of diversity. In particular, the other baselines, except for SAVI2I, cannot produce images of shorts or skirts. The StarGAN-v2 method can only synthesize trouser images and lacks shape diversity even for different inputs. On the contrary, SAVI2I produces results that are similar to its inputs. This may be ascribed to the fact that SAVI2I is more focused on continuous I2I translation so that the result of the synthesis can only exert influence over the input in color space. 2) For the metric of visual authenticity, Table \ref{upper2lower_stat} shows the FID of BC-GAN and other baselines on the `upper $\rightarrow$ lower' setting. We observe that our BC-GAN achieves the best score of visual authenticity in terms of FID. It outperforms the second-best method by a substantial margin: approximately 7.646. Among the baselines compared, MUNIT demonstrated the best performance in terms of FID. This can be attributed to its unique approach by employing separate style and content encoders, resulting in increased network capacity when compared to other alternatives. For qualitative observation, Fig. \ref{cmp_samples} (a) gives an illustration of synthesized images. It is clear that our method produces the most photo-realistic images in comparison to other methods. 3) For the metric of fashion compatibility, to reduce the subjectivity of the evaluation, F${\rm ^2}$BT is adopted as a metric to measure the results of BC-GAN and the baselines. As shown in Table \ref{upper2lower_stat}, BC-GAN delivers the best fashion compatibility score in terms of F${\rm ^2}$BT. Specifically, it achieves the best score with an advantage of 2.9\% compared with the second-best method. In Fig. \ref{cmp_samples} (a), the synthesized results of our model possess elements that harmonize with the given inputs. This indicates that our method synthesizes better results in terms of fashion compatibility. Meanwhile, although SAVI2I achieves the second-best fashion compatibility metric as shown in Table \ref{upper2lower_stat}, its synthesized images convey patterns that are too similar to its input as shown in Fig. \ref{cmp_samples} (a). On the contrary, BC-GAN has better texture generation, as its results harmonize more with their inputs, from the perspective of human observation. Taking the second sample for the `upper $\rightarrow$ lower' setting, we observe that BC-GAN produces lower clothing with vertical stripes which are compatible with the fashion style of the given upper clothing.

 \textbf{The `lower $\rightarrow$ upper' direction:} 1) For the metric of diversity, BC-GAN achieves the best LPIPS and has an improvement of approximately 0.058 in comparison with the second-best method as shown in Table \ref{lower2upper_stat}. Fig. \ref{cmp_samples} (b) also indicates that our method produces the best diversity among all methods. Taking the first sample for the `lower $\rightarrow$ upper' setting as an example, we can observe that BC-GAN produces the most different types of upper clothing. 2) For the metric of visual authenticity, BC-GAN achieves the best score in terms of FID and improves by 3.645 in comparison with the second-best method, as shown in Table \ref{lower2upper_stat}. Considering the visual comparison, the generated results listed in Fig. \ref{cmp_samples} (b) exhibit that BC-GAN synthesizes the most photo-realistic results among all methods. 3) For the fashion compatibility measurement, Table \ref{lower2upper_stat} illustrates that BC-GAN delivers the highest F${\rm ^2}$BT score among all methods. Moreover, Fig. \ref{cmp_samples} (b) also confirms that our method synthesizes the results with higher fashion compatibility. For example, it is observed that the synthetic clothing items are, with pink elements, compatible with the given clothing in the first sample for the `lower $\rightarrow$ upper' setting.
	
	\textbf{User Study}: In addition to the quantitative assessment with the above metrics, we undertook a user study for BC-GAN to delve deeper into the perceptual evaluation of the fashion compatibility exhibited by the generated clothing images listed in Table \ref{user_study}. From the DiverseOutfits dataset, we randomly selected 100 images for both the `upper $\rightarrow$ lower' and `lower $\rightarrow$ upper' translation directions as given clothing. Subsequently, we performed a user study to validate the superiority of our method. Our user study involved 10 participants who were not affiliated with the authors, yielding a total of 7,000 votes. Each participant was presented with three images: the original clothing, the clothing synthesized through our proposed method, and the clothing synthesized either by a comparable baseline or  from the DiverseOutfits dataset. They were instructed to select the outfit that displayed superior collocation. The results of the user study shown in Table \ref{user_study} demonstrate the substantial superiority of our method over the baseline approaches. It also indicates that the compatibility of fashion items synthesized by our method closely parallels the matching degree of expert-constructed outfits in the DiverseOutfits dataset.
	
	\begin{table*}[t]
		\footnotesize
		\centering
		\caption{User study results: Comparative preference rate of BC-GAN (ours) and baselines (Here, `Real' and `Random' refer to ground truth and randomly composed outfits from DiverseOutfits dataset, respectively)}
		\label{user_study}
		
		\begin{tabular}{p{0.85cm} P{2cm} P{2cm} P{2cm} P{2cm} P{2cm}  P{1.8cm} P{2cm}}
			\toprule
			& BC-GAN $>$ MUNIT & BC-GAN $>$ DRIT & BC-GAN $>$ DRIT++ & BC-GAN $>$ StarGAN-v2 & BC-GAN $>$ SAVI2I & BC-GAN $>$ Real & BC-GAN $>$ Random \\
			\hline
			Preference & 85.6 \%& 88.8\% & 91.8\% & 78.6\% & 95.5\% & 40.7\%& 76.7\% \\
			\bottomrule
		\end{tabular}
		\vspace{-0.4cm}
	\end{table*}

	\subsection{Ablation Study}
	
	This subsection presents the results of three sets of ablation studies carried out to validate the effectiveness of diversity loss, style embedding discriminator, and fashion compatibility discriminator.
	
	\begin{table}[t]
		\centering
		\caption{Comparison of BC-GAN with other variants without diversity loss in terms of diversity (LPIPS)}
		\label{abalation_per}
		\begin{tabular}{p{2.8cm} P{2.5cm} P{2.5cm}}
			\toprule
			Method & Setting & LPIPS($\uparrow$)\\
			\hline
			\multirow{2}{*}{BC-GAN w/o $\mathcal{L}_{div}$} & upper $\rightarrow$ lower & 0.418  \\
			& lower $\rightarrow$ upper & 0.527 \\
			\hline
			\multirow{2}{*}{BC-GAN (ours)} & upper $\rightarrow$ lower & \textbf{0.504}\\
			& lower $\rightarrow$ upper & \textbf{0.605}\\
			\bottomrule
		\end{tabular}
		\vspace{-0.4cm}
	\end{table}
	
	\begin{figure}[t]
		\centering
		\includegraphics[width=0.48\textwidth]{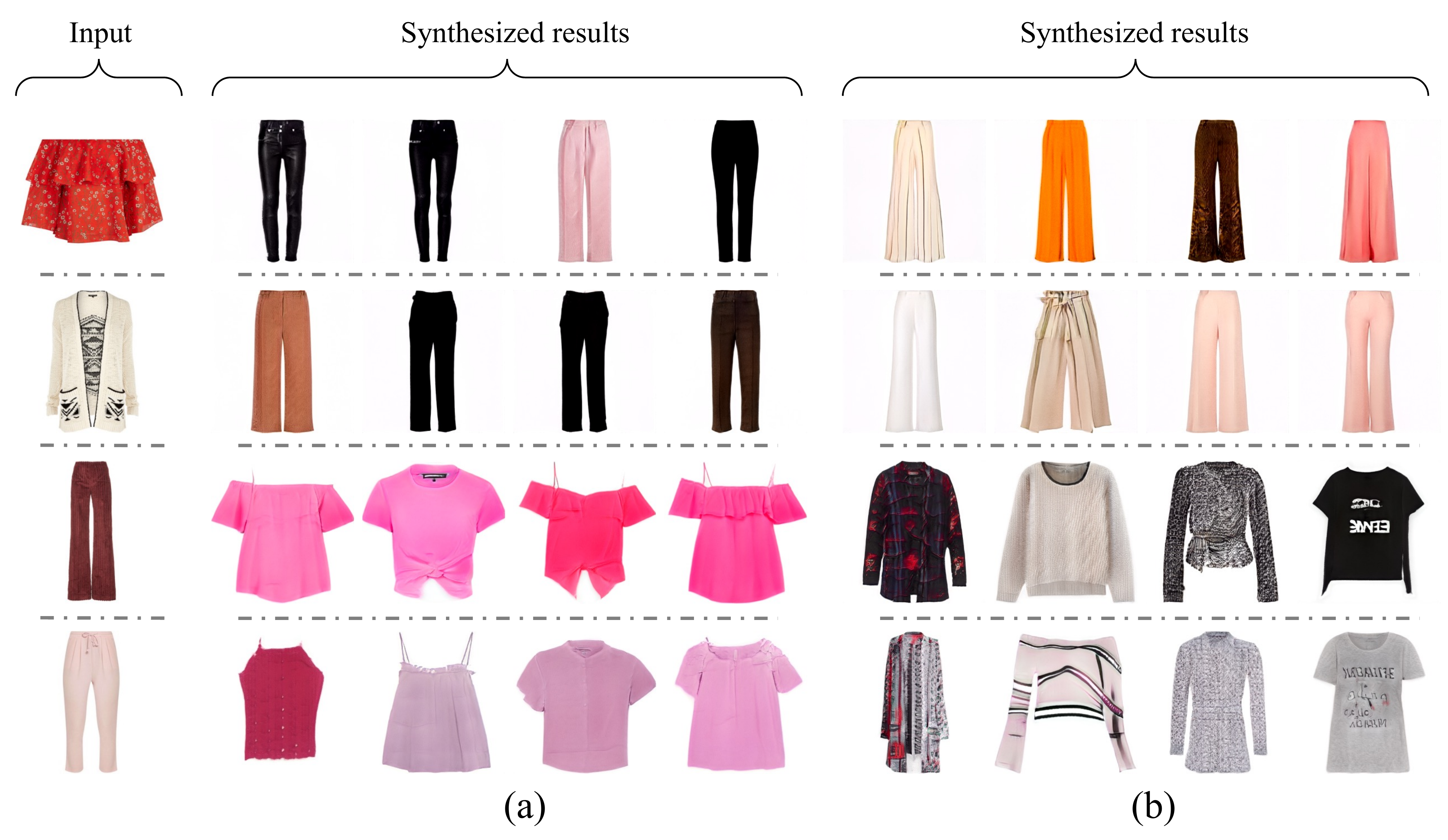}
		\caption{Visual comparison results for the effectiveness of diversity loss: (a) BC-GAN without diversity loss, and (b) BC-GAN with diversity loss (ours).}
		\label{cmp_samples_ds_fig}
		\vspace{-0.4cm}
	\end{figure}
	
	\textbf{Effectiveness of Diversity Loss}: The diversity loss is used to improve the diversity of synthetic results of BC-GAN. To investigate the effectiveness of diversity loss, we explore its impact on quantitative performance. The comparative results are given in Table \ref{abalation_per}. The results show that BC-GAN with diversity loss outperforms BC-GAN without diversity loss by a substantial margin: approximately 0.086 (or 0.078) on the `upper $\rightarrow$ lower' (or `lower $\rightarrow$ upper') translation setting. For qualitative comparison, we also provided certain synthetic samples in Fig. \ref{cmp_samples_ds_fig}. Taking the first and second columns of Fig. \ref{cmp_samples_ds_fig} as examples, we can see that BC-GAN without $\mathcal{L}_{div}$ only produces images of trousers for their given inputs. Moreover, the third and fourth columns of Fig. \ref{cmp_samples_ds_fig} show that the synthetic results deliver very similar colors. We believe this is caused by the fact that removing $\mathcal{L}_{div}$ degrades the variance of the synthetic samples in the target domain. This verifies that the addition of diversity loss benefits diversity in terms of both quantitative and qualitative evaluations.
	
		\begin{table}[t]
		\centering
		\caption{Comparison of BC-GAN with other variants without style embedding discriminator in terms of visual authenticity (FID)}
		\label{abalation_dse}
		\begin{tabular}{p{2.8cm} P{2.5cm} P{2.5cm}}
			\toprule
			Method & Setting & FID($\downarrow$)\\
			\hline
			\multirow{2}{*}{BC-GAN w/o $D_{se}$} & upper $\rightarrow$ lower & 339.445  \\
			& lower $\rightarrow$ upper & 331.327 \\
			\hline
			\multirow{2}{*}{BC-GAN w/ $D_{pix}$} & upper $\rightarrow$ lower &  291.862 \\
			& lower $\rightarrow$ upper & 260.163 \\
			\hline
			\multirow{2}{*}{BC-GAN (ours)} & upper $\rightarrow$ lower & \textbf{52.370}\\
			& lower $\rightarrow$ upper & \textbf{38.981}\\
			\bottomrule
		\end{tabular}
		\vspace{-0.4cm}
	\end{table}

\textbf{Effectiveness of Style Embedding Discriminator}:  We evaluated the effectiveness of the style embedding discriminator $D_{se}$ from two perspectives. Initially, we trained our BC-GAN without incorporating the $D_{se}$. As shown in Table \ref{abalation_dse}, the BC-GAN, lacking the style embedding discriminator, fails to ensure the realism of the synthesized clothing images. Subsequently, we trained our BC-GAN using a pixel-level discriminator instead of one in the $\mathcal{W}$ space. In Table \ref{abalation_dse}, `BC-GAN w/ $D_{pix}$' implies that the BC-GAN was trained with the original StyleGAN's real/fake discriminator, rather than the $D_{se}$. The results reveal that the BC-GAN with $D_{pix}$ is incapable of generating photo-realistic images, as evidenced by the high FID. This finding may seem counter-intuitive. However, upon analysis, it was deduced that the $D_{pix}$ supervises the generator's training by acting on the generated image, but its supervisory effect struggles to propagate to the encoding network $e$ over a long distance. Specifically, its pixel-level supervision is applied to the image generated by the synthesis network $g$. But, as the weights of the synthesis network are frozen during the BC-GAN training process, when its influence is back-propagated to the encoding network $e$, the weight update is hard to be achieved. In contrast, the style embedding discriminator $D_{se}$ proposed in this paper is designed to directly control the latent codes encoded from the encoding network $e$. Once the latent codes align with the distribution of the $\mathcal{W}$ space, the generated image can maintain a high level of visual fidelity.
		
\begin{figure}[t]
		\centering
		\includegraphics[width=0.48\textwidth]{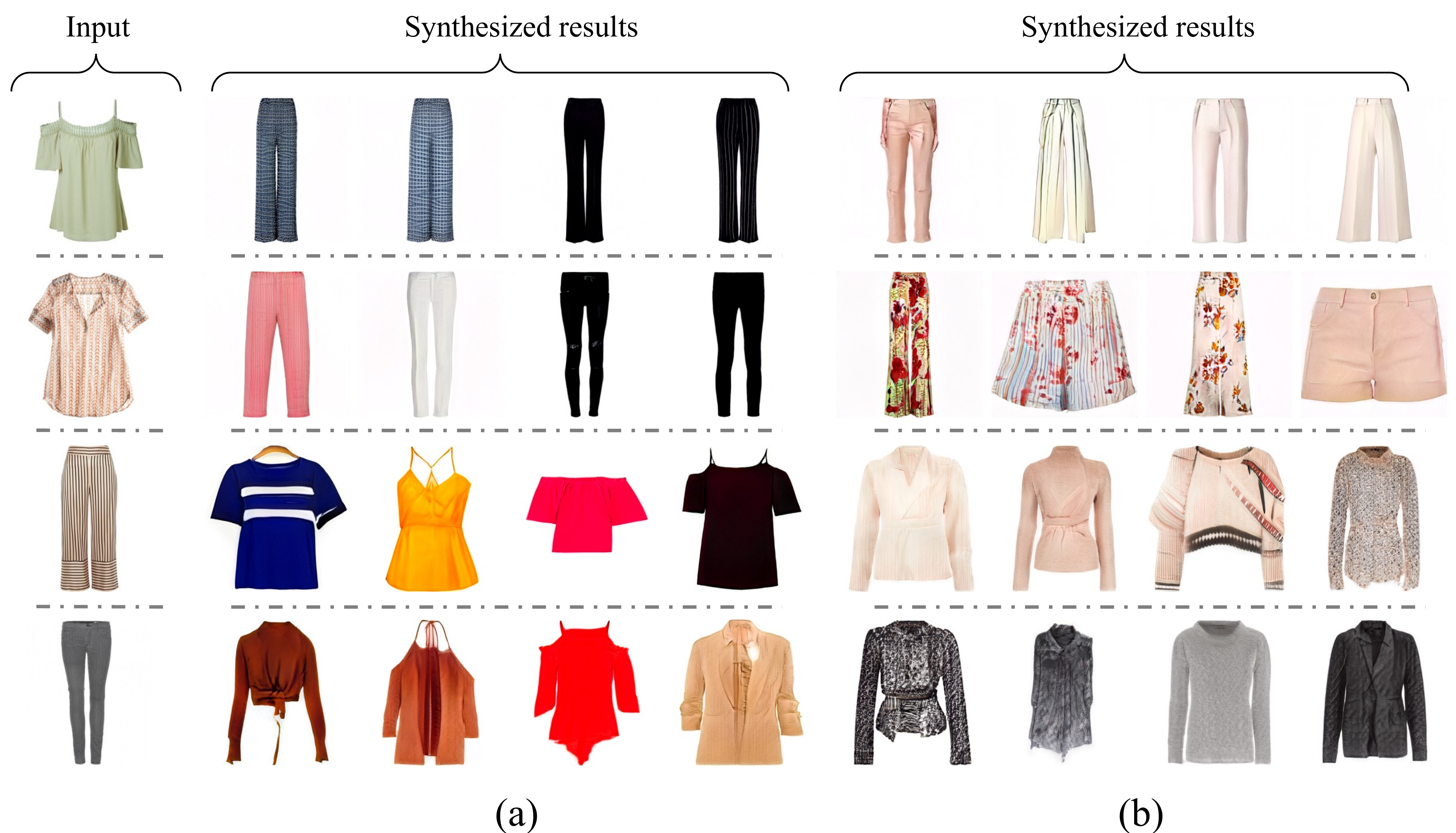}
		\caption{Visual comparison results for the effectiveness of compatibility discriminator: (a) BC-GAN without compatibility discriminator, and (b) BC-GAN with compatibility discriminator (ours).}
		\label{cmp_samples_cmp_fig}
		\vspace{-0.4cm}
	\end{figure}

	\textbf{Effectiveness of Compatibility Discriminator}: The compatibility discriminator is designed in a contrastive learning perspective to further improve the fashion compatibility in comparison to that of previous studies \cite{liu2019toward,liu2019collocating,yu2019personalized}. To investigate the effectiveness of our compatibility discriminator, we validate it from two aspects. For the first aspect, we first trained BC-GAN without using our compatibility discriminator. Then we compared this version with our original BC-GAN in terms of the aforementioned fashion compatibility metric, F${\rm ^2}$BT. As shown in Table \ref{abalation_cmp}, it is clear that our compatibility discriminator plays an important role in providing visually-collocated supervision on training BC-GAN by showing its contributions to the improvement in F${\rm ^2}$BT. We also generated some samples produced by BC-GAN with or without the use of the compatibility discriminator in Fig. \ref{cmp_samples_cmp_fig}. Obviously, synthetic clothing items harmonize better, which raises the fashion compatibility with the given clothing. For the second aspect, previous studies on collocated clothing generation only employed a discriminator to supervise the generator on real and fake pairs of given clothing and synthetic clothing. Here, we further compare our contrastive learning-based compatibility discriminator with the previous designs. In Table \ref{abalation_cmp}, `BC-GAN w/o contrastive learning' indicates that BC-GAN only takes real and fake pairs as its input. From Table \ref{abalation_cmp}, it is clear that BC-GAN can enhance its performance in terms of fashion compatibility by a significant amount.

	\begin{table}[t]
		\centering
		\caption{Comparison of BC-GAN with other variants without compatibility discriminator in terms of fashion compatibility (F${\rm ^2}$BT)}
		\label{abalation_cmp}
		\begin{tabular}{p{2.8cm} P{2.5cm} P{2.5cm}}
			\toprule
			Method & Setting & F$^ 2$BT ($\uparrow$)\\
			\hline
			\multirow{2}{*}{BC-GAN w/o ${\rm D}_{cmp}$} & upper $\rightarrow$ lower & 14.9\%\\
			& lower $\rightarrow$ upper & 17.0\%\\
			\hline
			\multirow{2}{*}{BC-GAN (ours)} & upper $\rightarrow$ lower & \textbf{22.9\%} \\
			& lower $\rightarrow$ upper & \textbf{22.8\%} \\
			\bottomrule
		\end{tabular}
	\end{table}
	
	\begin{table}[t]
		\centering
		\caption{Comparison of BC-GAN with other variants without contrastive learning in terms of fashion compatibility (F${\rm ^2}$BT)}
		\label{abalation_cmp_con}
		\begin{tabular}{p{2.8cm} P{2.5cm} P{2.5cm}}
			\toprule
			Method & Setting & F${\rm ^2}$BT ($\uparrow$)\\
			\hline
			\multirow{2}{*}{\makecell[l]{BC-GAN w/o contra-\\stive learning}} & upper $\rightarrow$ lower & 15.4\%\\
			& lower $\rightarrow$ upper & 17.8\%\\
			\hline
			\multirow{2}{*}{BC-GAN (ours)} & upper $\rightarrow$ lower & \textbf{22.9\%} \\
			& lower $\rightarrow$ upper & \textbf{22.8\%} \\
			\bottomrule
		\end{tabular}
		\vspace{-0.4cm}
	\end{table}
	
	\begin{figure*}[!ht]
		\centering
		\includegraphics[width=0.9\textwidth]{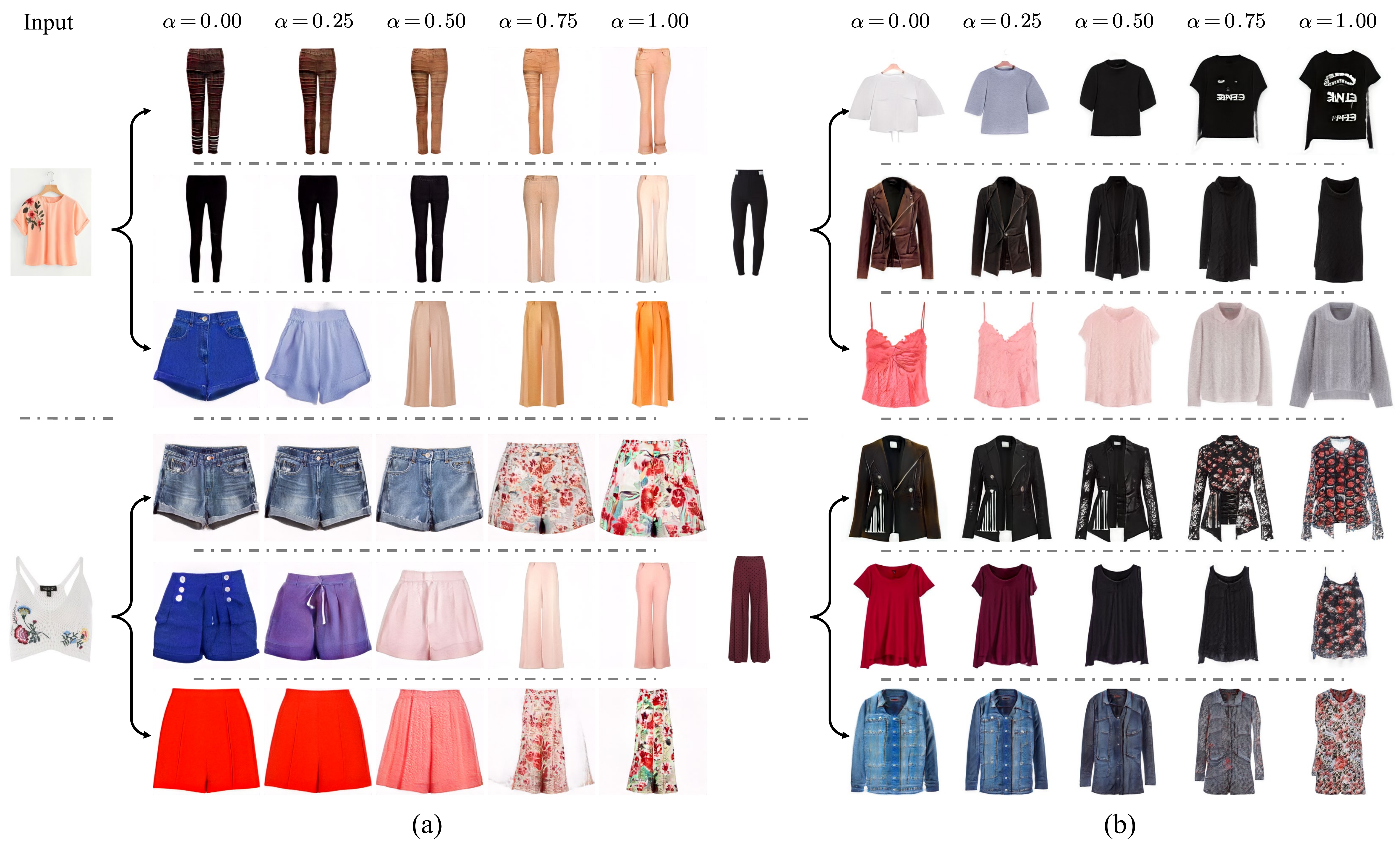}
		\caption{Linear interpolation between $\mathbf{w}_{i}^{orig}$ and $\mathbf{w}_{i}$, where $\mathbf{w}_{i}^{orig}$ is a style embedding mapped from a random latent code into $\mathcal{W}$ space, and $\mathbf{w}_{i}$ is a style embedding encoded from a given clothing image and a random synthesized image into $\mathcal{W}$ space: (a) continuous translation on the lower domain, and (b) continuous translation on the upper domain.}
		\label{morph_samples}
		\vspace{-0.4cm}
	\end{figure*}

	\begin{table}[t]
		\centering
		\caption{The times of better fashion compatibility (F${\rm ^2}$BT) when the mixed style embedding is provided a larger mix ratio}
		\label{cont_trans_tab}
		\begin{tabular}{p{1.8cm} P{2.8cm} P{2.8cm} }
			\toprule
			\multirow{3}{*}{Mix ratio $\alpha$} & \multicolumn{2}{c}{Setting}\\
			\cmidrule[0.5pt]{2-3}
			& upper $\rightarrow$ lower & lower $\rightarrow$ upper\\
			\hline
			0.00 vs 0.25 & 50.5\% & 49.3\%\\
			0.25 vs 0.50 & 53.4\% & 52.0\%\\
			0.50 vs 0.75 & 54.4\% & 53.1\%\\
			0.75 vs 1.00 & 52.8\% & 51.1\%\\
			\bottomrule
		\end{tabular}
		\vspace{-0.4cm}
	\end{table}

	\subsection{Additional Study}
	\label{additional_study}
	
	Our BC-GAN is designed with the idea of ``encoding random images with the given image first, decoding them for better fashion compatibility later.'' Attributable to the disentanglement of the learned $\mathcal{W}$ space, we can easily perform continuous translation on the target domain. In this subsection, we conduct an additional study to validate whether BC-GAN works under the above-mentioned mechanism. In previous research \cite{mao2022continuous}, continuous I2I translation aims at morphing images from the source domain $\mathcal{X}$ into the target domain $\mathcal{Y}$. However, continuous I2I translation between upper clothing and lower clothing makes no sense in our research problem. In real scenarios, users do not need an image looking both like the upper and the lower clothing. On the contrary, by taking advantage of the disentangled features in the $\mathcal{W}$ space, our framework can support I2I translation from a randomly sampled synthetic clothing item to a new clothing item which is more compatible with a given clothing item from a visual perspective. Specifically, the continuous translation on the target domain is a linear interpolation based on the style embedding. Formally, it can be formulated as follows:
	\begin{equation}
		\mathbf{w}_{i}^{mix} = \mathbf{w}_{i}^{orig} \times (1 - \alpha) + \mathbf{w}_{i} \times \alpha,
	\end{equation}
	where $\mathbf{w}_{i}^{mix}$ is a style embedding mixed from $\mathbf{w}_{i}^{orig}$ and $\mathbf{w}_{i}$ with the ratio $\alpha$. The mixed style embedding $\mathbf{w}_{i}^{mix}$ is fed into the pre-trained synthesis network $g$ to obtain an RGB image. Here, we compare the fashion compatibilities coming from different mix ratios of random style embeddings and encoded style embeddings. As shown in Table \ref{cont_trans_tab}, we compare different mix ratios in the style embeddings. We adopt five different mix ratios in $\{0.00, 0.25, 0.50, 0.75, 1.00\}$ to conduct experiments in terms of fashion compatibility. For $0.00$ vs $0.25$, we count the number of times that the synthetic results with a ratio of $0.25$ are better than that of $0.00$ in terms of fashion compatibility. Here, the fashion compatibility comparison used the fashion compatibility predictor $\varphi$ mentioned in Section \ref{eva_metric}. For example, the first row of Table \ref{cont_trans_tab} on `upper $\rightarrow$ lower' setting means the synthesized results using a mix ratio of 0.25 beat the results of the mix ratio of 0.00 over 50.5\% of the time. As shown in Table \ref{cont_trans_tab}, we can observe that a higher mix ratio is able to improve the fashion compatibility of the outfits. Meanwhile, we see that 0.00 vs 0.25 on `lower $\rightarrow$ upper' has a decrease in terms of fashion compatibility. The worse performance may stem from the fact that the conversion from $\mathbf{w}_{i}^{orig}$ to $\mathbf{w}_{i}$ cannot be easily disentangled for a high-level learning, i.e., fashion compatibility learning. Moreover, it is observed that the fashion compatibility of the synthesized samples improves with an increase of mix ratio, as shown in Fig. \ref{morph_samples}. Taking the first given upper clothing as an example, the synthesized results become more and more compatible with an increase of the mix ratio. This indicates that despite taking a random synthesized image to produce diverse images, BC-GAN is able to encode random information into valid samples and improve the fashion compatibility of the synthesized clothing.

	\section{Conclusion and Future Work}
	\label{conclusion}
	In this paper, we proposed a new collocated clothing generation framework by leveraging the power of a pre-trained model. Our framework can synthesize multiple visually-collocated and diverse items of clothing based on the given clothing at the same time. It is composed of a generator to produce targeted images, a style embedding discriminator to ensure the encoded style embedding fits the distribution of latent space of the used pre-trained model, and a compatibility discriminator to supervise the generator capable of synthesizing visually-collocated clothing images. Experiments on a large-scale dataset constructed by ourselves show that the proposed method outperforms alternative methods in terms of diversity, visual authenticity, and fashion compatibility. In future work, it would be interesting to synthesize diversified fashion items in multiple fashion categories in a more controllable manner. In addition, diffusion probabilistic models \cite{ho2020denoising,rombach2022high} have been experiencing a surge in interest due to their training stability and their promising results on text-to-image generation. In spite of this, GANs are currently faster in synthesizing images and use less memory when compared to diffusion models during the training phase. In the future, developing collocated clothing synthesis based on advanced diffusion models driven by multi-modal information deserves further investigation.
	
	\bibliographystyle{IEEEtran}
	\bibliography{ref}

\end{document}